\documentclass[sigconf]{acmart}
\AtBeginDocument{%
  }

\copyrightyear{2026}
\acmYear{2026}
\setcopyright{cc}
\setcctype{by}
\acmConference[MM '26]{Proceedings of the 34th ACM International Conference on Multimedia}{November 10--14, 2026}{Rio de Janeiro, Brazil}
\acmBooktitle{Proceedings of the 34th ACM International Conference on Multimedia (MM '26), November 10--14, 2026, Rio de Janeiro, Brazil}
\acmDOI{10.1145/3767308.3835888}
\acmISBN{979-8-4007-2213-4/2026/11}
\usepackage{algorithm}
\usepackage{algorithmic}
\usepackage{pifont}        
\usepackage{fontawesome5}  
\usepackage{xcolor}
\usepackage{graphicx}
\usepackage{makecell}
\usepackage{multirow}
\usepackage{longtable}
\usepackage{array}
\usepackage{balance}

\newcommand{\Yes}{\textcolor{green!60!black}{\ding{51}}} 
\newcommand{\No}{\textcolor{red!70!black}{\ding{55}}} 
\newcommand{\Partial}{\textcolor{orange!80!black}{\faExclamationTriangle}} 

\begin{document}

\title{WARP: A Unified Benchmark for Invisible Image Watermarking — Robustness and Protection Against Attacks}

\author{Khaled Abud}
\affiliation{%
  \institution{MSU Institute for Artificial Intelligence}
  \city{Moscow}
  \country{Russia}
}
\authornote{
 Corresponding Authors, Emails: khaled.abud@graphics.cs.msu.ru,\\ yakushev@ispras.ru, a.akimenkov@ispras.ru
}

\author{Aleksey Yakushev}
\authornotemark[1]
\affiliation{%
  \institution{Trusted AI Research Center RAS}
  \city{Moscow}
  \country{Russia}}

\author{Aleksandr Akimenkov}
\authornotemark[1]
\affiliation{%
  \institution{Trusted AI Research Center RAS}
  \city{Moscow}
  \country{Russia}
}

\author{Irina Serzhenko}
\affiliation{%
 \institution{Independent researcher}
 \city{Moscow}
 \country{Russia}}

\author{Kirill Aistov}
\affiliation{%
  \institution{MSU Institute for Artificial Intelligence}
  \city{Moscow}
  \country{Russia}}

\author{Egor Kovalev}
\affiliation{%
  \institution{MSU Institute for Artificial Intelligence}
  \city{Moscow}
  \country{Russia}}

\author{Dmitry Obydenkov}
\affiliation{%
  \institution{Trusted AI Research Center RAS}
  \city{Moscow}
  \country{Russia}}

\author{Sergey Lavrushkin}
\affiliation{%
  \institution{MSU Institute for Artificial Intelligence}
  \city{Moscow}
  \country{Russia}}

\author{Anastasia Antsiferova}
\affiliation{%
  \institution{Trusted AI Research Center RAS}
  \city{Moscow}
  \country{Russia}}

\author{Dmitriy Vatolin}
\affiliation{%
  \institution{MSU Institute for Artificial Intelligence}
  \city{Moscow}
  \country{Russia}}

\author{Yury Markin}
\affiliation{%
  \institution{Trusted AI Research Center RAS}
  \city{Moscow}
  \country{Russia}}

\author{Kirill Lukianov}
\affiliation{%
  \institution{Trusted AI Research Center RAS}
  \city{Moscow}
  \country{Russia}}

\renewcommand{\shortauthors}{Khaled Abud et al.}

\begin{abstract}
Digital image watermarking is increasingly critical in media contexts, as emerging regulations and industry practices require marking AI-generated content and ensuring traceable sources to prevent manipulation or misuse. 
Recent advances in invisible watermarking methods highlight the need to update existing benchmarking practices to reflect current techniques and evaluation criteria.

We address this by introducing \textbf{WARP} — a unified framework and benchmark for evaluating the robustness of invisible watermarks. WARP incorporates 32 recent classical, deep, and generative watermarking methods, as well as 34 different erasing techniques, ranging from traditional distortions to more sophisticated adversarial, purification, and re-embedding attacks. It provides standardized, reproducible, and easily scalable protocols for evaluating perceptual quality, watermark readability, and attack resilience.

Using WARP, we extensively evaluate current invisible watermarking techniques, collecting \textit{the largest robustness benchmark in the field}. Results identify the most robust approaches under both distortion and adversarial conditions, and reveal consistent relationships between watermarking methods and the attack strategies most effective against them. Our experiments also highlight that some of the watermarking methods considered are highly vulnerable to reembedding, even if they are robust to standard distortions. The code is made available at \url{https://github.com/ispras/wibe}.

\end{abstract}




\maketitle

\section{Introduction}
\begin{figure*}[htbp]
    \centerline{\includegraphics[width=0.99\textwidth]{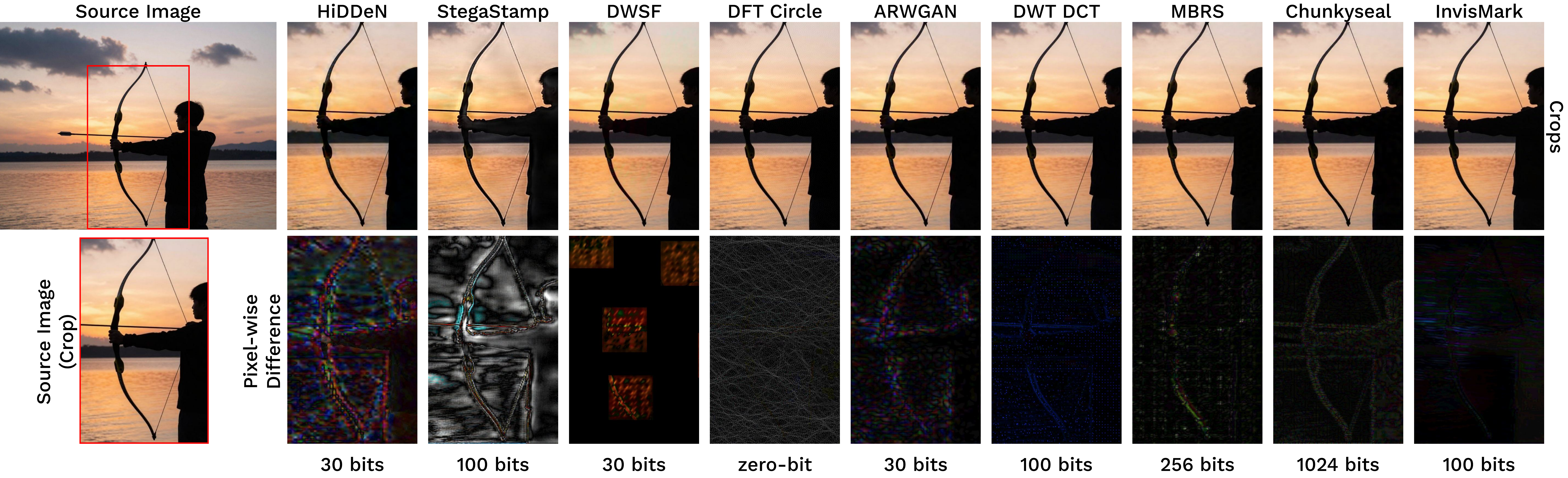}}
    \caption{Visualizations of multiple invisible watermarking methods available in our benchmark. The second row shows the pixel-wise difference from the source image (magnified $\times7$ for visibility) alongside the embedded message capacity in bits.}
    \Description{A grid of image thumbnails. The top row shows a source photograph of an archer silhouetted against a sunset, followed by its watermarked versions produced by nine methods: HiDDeN, StegaStamp, DWSF, DFT Circle, ARWGAN, DWT DCT, MBRS, ChunkySeal and InvisMark. The bottom row shows the corresponding pixel-wise differences from the source, magnified seven times, with each method's message capacity printed underneath, ranging from zero-bit to 1024 bits. All watermarked images are visually indistinguishable from the source, while the difference maps reveal method-specific signatures: dense high-amplitude texture for HiDDeN and StegaStamp, block and ring structures for the frequency-domain methods, and almost entirely black maps for ChunkySeal and InvisMark.}
\label{fig:viz}
\end{figure*}

The rapid increase in generative and AI-produced images in digital media has created an urgent need for mechanisms that ensure content provenance, authenticity, and traceability. Digital watermarking has emerged as a primary technology for these purposes, enabling imperceptible embedding of ownership or source information directly into an image. Beyond copyright protection, watermarking plays a critical role in mitigating misinformation, supporting journalistic verification, and complying with emerging regulatory frameworks that require explicit marking of AI-generated content. 
While recent deep learning and generative watermarking methods have improved both imperceptibility and capacity (Fig.~\ref{fig:viz}), their robustness evaluations often remain fragmented and underexplored. This creates a gap between laboratory performance and the practical reliability required for real-world media applications.

Several limitations in current watermarking benchmarks hinder systematic assessment of modern methods. First, most publicly available benchmarks~\cite{kutter1999fair, an2024waves} were developed two or more years ago and do not incorporate state-of-the-art deep and generative techniques. Second, evaluations are typically restricted to simple distortion models, ignoring complex attack scenarios such as adaptive removal strategies \cite{alam2025saliency, souvcek2025transferable, mullerBlackBoxForgeryAttacks2025} or competitive re-embedding, where a watermark is intentionally overwritten by another. Third, existing frameworks rarely consider the interplay between perceptual quality, robustness, and resistance to overwriting in a unified evaluation, making it difficult to compare methods across diverse threat models or to identify systematic vulnerabilities. Consequently, there is no standardized methodology that captures the multifaceted nature of watermark performance in contemporary media contexts.

The goal of this work is to develop a comprehensive, reproducible benchmark for invisible image watermarking that standardizes the assessment of both perceptual quality and robustness across modern techniques. In addition, the study aims to systematically evaluate the current state of the field by assembling a representative set of methods that collectively cover the full spectrum of existing approaches to both watermarking and corresponding attack strategies. Specifically, this study addresses three core research questions:

\begin{enumerate}
\renewcommand{\labelenumi}{RQ\arabic{enumi}}
    \item Can any watermarks survive both standard perturbations, erasing attacks, and cross-domain adversarial purification and re-generation techniques?
    \item What systematic failure modes exist within specific method classes when exposed to state-of-the-art removal attacks?
    \item To what extent can an existing watermark be overwritten via competitive re-embedding without compromising the carrier image?
\end{enumerate}

These questions aim to cover currently underexplored robustness to complex state-of-the-art watermark removal attacks, which will enable finding watermarking methods most suitable for practical scenarios, providing actionable insights for both researchers and practitioners in media protection and content provenance.

We summarize our contributions as follows:

\begin{itemize}
    \item \textbf{Comprehensive benchmark}. We conduct a large-scale evaluation of 32 classical, deep, and generative watermarking methods across 34 adversarial scenarios that reflect all state-of-the-art practices at the time of submission.
    \item \textbf{Scalable evaluation framework}. All experiments are carried out using our open-source framework designed for modularity and reproducibility. To the best of our knowledge, WARP implements the largest number of watermarking methods and attacks available in the field.
    \item \textbf{Practical insights}. Our results provide valuable recommendations for selecting watermarking strategies based on method-specific strengths and vulnerabilities, facilitating informed deployment in real-world media systems.
    \item \textbf{Evaluation of watermark robustness to overwriting}. A novel methodology to systematically assess the potential for competitive re-embedding across watermarks, addressing a gap unexamined in prior work.

\end{itemize}

\section{Related Work}
\subsection{Invisible Watermarking Methods}

\label{sec:wm_methods}
Watermarking techniques can be classified by the stage at which the watermark is embedded. \textit{Generative} (built-in) methods integrate it directly into the image creation process, e.g., within diffusion-based generative models, whereas \textit{post-hoc} methods add it to already existing images through transformations in spatial, frequency, or latent domains. The former are well-suited to AI-generated content provenance and are inherently robust to generation-specific distortions; the latter offer flexibility for legacy or arbitrary images but may require additional steps to maintain imperceptibility.

\begin{table*}[h]
\caption{Comparison of frameworks for watermarking research.  "Benchmark" column indicates whether the framework includes comprehensive results for the implemented methods, with \Partial $\:$ denoting limited evaluations. * next to the number denotes that the count is based on the methods reported in the original paper, but the number of fully implemented watermarking pipelines found in public repository is smaller. We consider methods recent if they have been released after Jan 2025.}

\centering

\begin{tabular}{lccccccc}
\toprule
\multirow{2}{*}{\textbf{Name}} 
& \multicolumn{4}{c}{\textbf{WA Methods}} 
& \multirow{2}{*}{\textbf{Attacks}} 
& \multirow{2}{*}{\textbf{Metrics}} 
& \multirow{2}{*}{\textbf{Benchmark}} \\
& \textbf{Post-hoc} & \textbf{Built-in} & \textbf{Total} & \textbf{Recent methods}  & & & \\
\midrule
Stirmark \cite{kutter1999fair} & 0 & 0 & 0 & 0 & 20 & 4 & \Partial \\
invisible-watermark \cite{InvisibleWatermark} & 3 & 0 & 3 & 0 & 0 & 1  & \No \\
SSL watermarking \cite{fernandez2022sslwatermarking} & 2 & 0 & 2 & 0 & 6 & 4 & \No \\
WAVES \cite{an2024waves} & \(3^*\) & \(2^*\) & 5 & 0 & 20 & 9 & \Yes \\
W-Bench \cite{luRobustWatermarkingUsing2025} & 11 & 0 & 11 & 2 & 13 & 7 & \Yes \\
MarkDiffusion~\cite{pan2025markdiffusion} & 0 & \textbf{\(9^*\)} & \(9^*\) & 5 & \(9^*\) & \(12^*\) & \Partial \\
\midrule
WIBE v1.0~\cite{yakushev2025wibe}  & 15 & 2 & 17 & 2 & 23 & 10 & \Partial\\
\textbf{WARP (a.k.a. WIBE v2.0)} & \textbf{26} & 6 & \textbf{32} & \textbf{9} & \textbf{34} & \textbf{14} & \textbf{\Yes} \\
\bottomrule
\end{tabular}
\label{tab:wm_tools}
\end{table*}

\textbf{Generative Watermarking Methods}. 
Most in-generation approaches modify the initial noise of the diffusion models. 
TreeRing~\cite{wen2023tree} embeds a structured pattern in the frequency domain of the initial noise to ensure robustness to geometric distortions. 
Its extensions~\cite{ci2024ringid, varlamov2024metr} use discretized ring patterns to enable multi-bit watermarking. 
MaXsive~\cite{mao2025maxsive} separates functions: an X-shaped template corrects geometric distortions, while a separate channel carries the message. 
Gaussian Shading~\cite{yang2024gaussian} preserves the original noise distribution and encodes bits using quantiles of the normal distribution.
An alternative direction, implemented in Stable Signature~\cite{fernandez2023stable}, is to integrate watermarks by partly fine-tuning the generation model (e.g., Stable Diffusion~\cite{rombach2022high}).

\textbf{Post-hoc methods}, applicable to already existing images, encompass both classical frequency-domain and modern neural architectures. 
Traditional approaches usually operate in Discrete Wavelet~\cite{islam2020svm}, Fourier~\cite{poljicak2011discrete}, or Cosine~\cite{al2007combined} Transform domains and embed watermarks in selected frequency coefficients.
More advanced methods~\cite{guzik2015real} utilize Spread Spectrum techniques like CAISS~\cite{valizadeh2010correlation}. 

Starting with HiDDeN~\cite{zhu2018hidden}, the "encoder --- noise layer --- decoder" architecture dominates post-processing watermarking.
\cite{ma2022towards, fang2023flow} utilize invertible networks to improve accuracy, while~\cite{jia2021mbrs} solves JPEG non-differentiability.
To enhance imperceptibility, \cite{huang2023arwgan, zhang2019robust, fernandez2024video, souvcek2025pixel} apply attention mechanisms and adversarial losses, and \cite{bui2023trustmark, xu2025invismark} leverage difference scaling for high resolution. 
For resilience against geometric attacks, \cite{guo2023practical, sander2024watermark, hu2025mask} implement multi-region marking and masking, whereas SyncSeal~\cite{fernandez2025geometric} focuses on synchronization.
SSL~\cite{fernandez2022sslwatermarking} inherits robustness from the latent space invariance of the DINO~\cite{caron2021emerging} model.
Robust-Wide~\cite{hu2024robust} protects against instruction-based editing.
StegaStamp~\cite{tancik2020stegastamp} and PIMoG~\cite{fang2022pimog} apply more aggressive perturbations during training to address physical print-cam and screen-cam scenarios.
ChunkySeal~\cite{petrov2025we} demonstrates the potential to significantly increase capacity without reducing the robustness.

These methods collectively represent the current state of the art in image watermarking, spanning zero-bit to multi-bit scenarios, frequency-domain classics to fully learned pipelines, and in-generation to post-generation paradigms. The present benchmark evaluates their performance under a unified set of removal and perturbation attacks, highlighting relative strengths and vulnerabilities in real-world robustness scenarios.

\subsection{Watermark Robustness Evaluation}

Watermarking research is inherently tied to robustness assessment, as adversaries are incentivized to remove existing watermarks or forge them on unprocessed images. Work on erasing watermarks with minimal quality loss, in turn, drives the development of algorithms designed to resist such countermeasures.

Resistance to image transformations that might occur naturally is expected from most watermarking methods. These include image compression, cropping, resizing, color manipulation, and noise.
Recent research also suggests that some watermarks can be removed by relatively simple image manipulation. Yang et al. \cite{yangSteganalysisDigitalWatermarking2024} propose extracting the watermark pattern by averaging several marked images and subtracting the resulting pattern. Shamshad et al. \cite{shamshadFirstPlaceSolutionNeurIPS2025} suggests that shifting the image by several pixels horizontally or vertically can remove watermarks such as TreeRing. In addition to natural image distortions, more targeted attacks on watermarking methods have also been investigated.
Some watermarks are shown to be vulnerable to image regeneration using diffusion models~\cite{saberiRobustnessAIImageDetectors2023, zhaoInvisibleImageWatermarks2024, liuImageWatermarksAre2025}, VAEs~\cite{balle2018variational} and Deep Image Prior~\cite{liangBaselineMethodRemoving2025}.
Another category of attacks is based on adversarial optimization techniques~\cite{lukasLeveragingOptimizationAdaptive2023, saberiRobustnessAIImageDetectors2023, mullerBlackBoxForgeryAttacks2025}.

\begin{figure*}[htbp]
    \centerline{\includegraphics[width=0.99\textwidth]{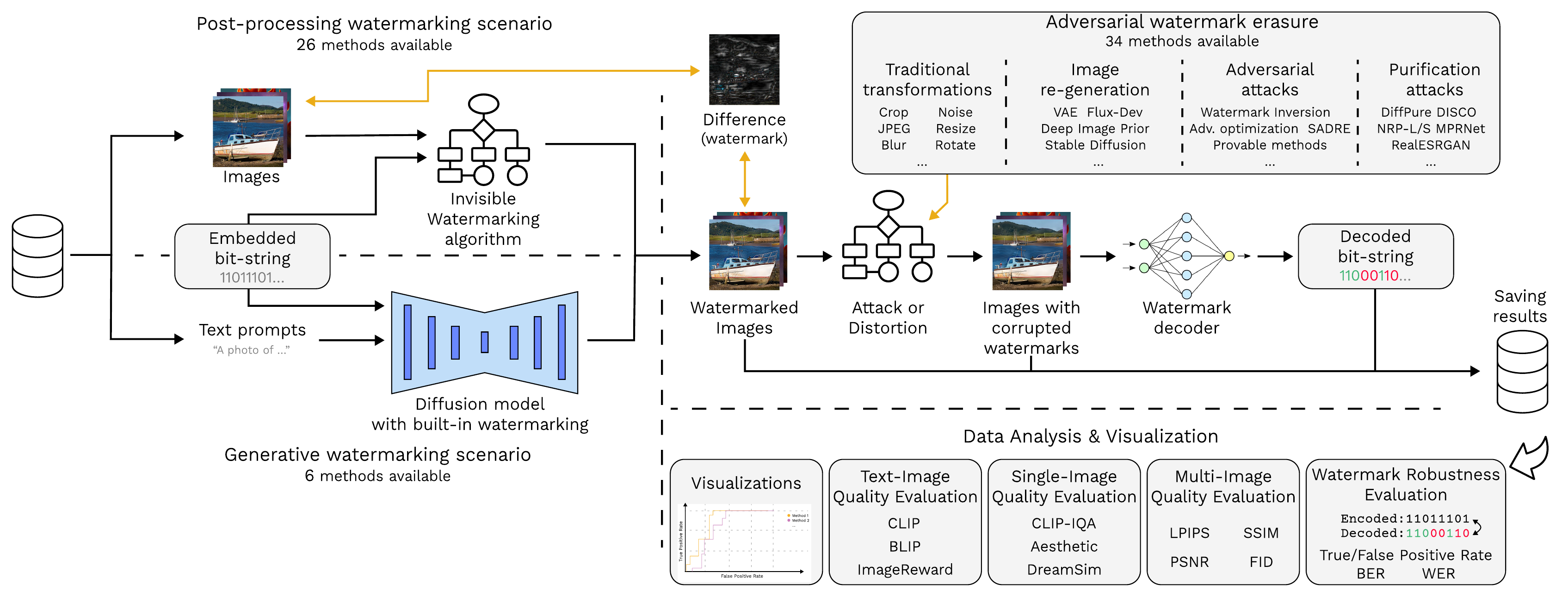}}
    \caption{Overview of the WARP framework pipeline, inherited from WIBE~\cite{yakushev2025wibe}. The process begins with (1) Watermark Embedding and (2) Quality \& Imperceptibility Evaluation on the left. The central module, (3) Attack Simulation, applies diverse perturbations or adversarial transformations. On the right, the pipeline continues with (4) Watermark Extraction, (5) Post-Attack Evaluation, and (6) Aggregation \& Logging, while (7) Visualization \& Reporting is performed as the final step.}
    \Description{A block diagram of the WARP evaluation pipeline, flowing from left to right. On the left, source images together with an embedded bit-string enter either a post-hoc invisible watermarking algorithm, of which 26 are available, or a diffusion model with built-in watermarking driven by text prompts, of which 6 are available; both produce watermarked images, and the difference from the source feeds a quality and imperceptibility evaluation. In the centre, an attack module applies one of the available erasure techniques, grouped into traditional transformations such as crop, JPEG, blur, noise, resize and rotate; image re-generation such as VAE, Deep Image Prior and Stable Diffusion; adversarial attacks such as watermark inversion and SADRE; and purification attacks such as DiffPure, DISCO, NRP and RealESRGAN. The result is images with corrupted watermarks. On the right, a watermark decoder extracts a bit-string that is compared with the embedded one, and results are logged and passed to a data analysis and visualisation stage covering text-image, single-image and multi-image quality metrics and watermark robustness metrics including BER, WER and true and false positive rates.}
\label{fig:main-pipeline}
\end{figure*}

Several comprehensive benchmarks have been proposed to rigorously evaluate watermark robustness.
Kutter and Petitcolas~\cite{kutter1999fair} propose an early benchmark for image watermarking that standardizes evaluation by defining a set of common 
attack types (e.g., compression, geometric transformations, filtering, and noise addition) and emphasizing joint assessment of robustness and perceptual quality. The framework relies on a limited set of core evaluation metrics, primarily including bit-error rate for robustness, perceptual quality measures aligned with the human visual system, and ROC-based detection analysis. However, it does not implement or benchmark specific watermarking methods, instead focusing on defining a fair and reproducible evaluation protocol.
WAVES~\cite{an2024waves} was the first substantial benchmark, featuring 20 attacks based on image distortion, regeneration, and adversarial attacks; however, only 3 watermarks were evaluated.
W-Bench~\cite{luRobustWatermarkingUsing2025} evaluates 11 watermarking methods against 13 attacks including image distortion, regeneration, global and local editing, and image-to-video generation.
MarkDiffusion~\cite{pan2025markdiffusion} is an open-source toolkit for generative watermarking of latent diffusion models, providing a unified implementation of image watermarking methods, along with visualization and evaluation modules. In the image domain, it integrates 9 watermarking algorithms and supports a comprehensive evaluation pipeline with 9 attack types (e.g., geometric transformations, noise, and compression) and 12 evaluation metrics for assessing detectability and visual quality. However, as a toolkit, it primarily focuses on standardization and usability rather than conducting a large-scale comparative benchmark across methods. 
WIBE~\cite{yakushev2025wibe} is our evaluation framework focusing on modularity and extensibility, but it does not provide detailed benchmarking results, as it was published as a tool demonstration. In this work, we use WIBE as a starting point: we significantly extend its contents to include all recent state-of-the-art watermarks, attacks and evaluation metrics, and further refine its architecture. We then employ our library to conduct large-scale evaluations and release WARP, a comprehensive watermark robustness benchmark. 
As summarized in Table~\ref{tab:wm_tools}, our framework comprises the largest number of watermarking and erasing methods to date and extensively tests them to provide detailed insights into the current state of watermark robustness.

\section{Robustness Evaluation Framework}

To ensure consistent and reproducible experimental conditions, we introduce \textbf{WARP}, a modular framework for evaluating the robustness of invisible image watermarking methods. An overview of the pipeline is shown in Fig.~\ref{fig:main-pipeline}. 
WARP adopts a plugin-based architecture with YAML-driven experiment configuration, enabling flexible and transparent setup of evaluation protocols. It supports multiple data formats and storage backends, provides real-time visual feedback, and can orchestrate experiments across multi-GPU systems. The framework operates within isolated Python environments and is capable of automatically resolving method-specific dependencies.

WARP inherits the overall design and structure of WIBE, but significantly expands its functionality, scope, and scale. Architecturally, WARP introduces a uv-based automatic dependency resolver that builds and separates the environments for incompatible methods, as well as additional metrics and other novel features like re-embedding and multi-step attack setups. The evaluation workflow is structured as a sequence of configurable processing stages, allowing watermark embedding, quality assessment, attack application, watermark extraction, and result logging to be implemented as independent components. This design facilitates rapid experimentation and fair comparison across diverse methods.

To the best of our knowledge, WARP integrates the largest collection of image watermarking and watermark removal techniques within a unified framework. At the same time, new watermarks, attacks, metrics, and datasets can be easily incorporated by wrapping their structure with WARP’s lightweight API.

\section{Benchmark}

\label{sec:bench}
\noindent\textbf{Watermarking methods}. To cover the field of invisible watermarking, we include a diverse set of 26 post-hoc and 6 generative (built-in) watermarks in our study. Table 2 
in the Appendix summarizes all evaluated methods. Architecturally, tested methods range from classical algorithms like \textbf{DWT DCT}~\cite{al2007combined}, \textbf{DCT CAISS}~\cite{guzik2015real} up to the most recent models that employ complex adversarial training pipelines (e.g., \textbf{PixelSeal}~\cite{souvcek2025pixel}, \textbf{Robust-Wide}~\cite{hu2024robust}). In terms of capacity, they cover a full spectrum from zero-bit methods that rely on statistical tests to infer a binary result   (e.g., \textbf{DFT Circle}~\cite{poljicak2011discrete}) to large multi-bit watermarks (e.g., 1024 bits in \textbf{ChunkySeal}~\cite{petrov2025we}) that can hide large messages and double as steganography methods. Generative watermarks include common \textbf{Stable Signature}~\cite{fernandez2023stable} and \textbf{TreeRing}~\cite{wen2023tree} algorithms, their modifications (\textbf{RingID}~\cite{ci2024ringid}, \textbf{METR}~\cite{varlamov2024metr}), and more recent methods like \textbf{MaxSive}~\cite{mao2025maxsive}.

\noindent\textbf{Erasure methods}. To thoroughly investigate the robustness of digital watermarks, we collect a diverse set of image processing methods that could be utilized to evade watermark recognition. The full list of employed methods alongside their brief descriptions can be found in Table 4 
in the Appendix. We broadly categorize the attacks into 4 distinct groups: traditional distortions, re-generation attacks, adversarial purification methods and erasure attacks. Traditional distortions include classical transformations that are often used to test watermarks, such as \textbf{JPEG compression}, \textbf{noise}, \textbf{crops} and \textbf{blur}. Re-generation attacks utilize pretrained generative models (e.g., \textbf{VAEs}~\cite{balle2018variational}, \textbf{Stable Diffusion}~\cite{rombach2022high}, \textbf{FLUX}~\cite{flux2024}) to alter the watermarked image, often resulting in significant changes relative to the source. Adversarial purification methods incorporate various techniques used to smooth out the image and remove potential adversarial perturbations embedded into the image. These methods are typically employed as \textbf{adversarial defenses} against attacks in other computer vision domains~\cite{gushchin2024guardians}. The erasure attacks group consists of image processing methods that were deliberately proposed or tested as attacks against invisible watermarks in the literature. Representative methods include \textbf{WMForger}~\cite{souvcek2025transferable}, \textbf{DIP~\cite{liangBaselineMethodRemoving2025}}, \textbf{Adversarial Embedding}~\cite{an2024waves} and \textbf{Averaging} attack~\cite{yangSteganalysisDigitalWatermarking2024}.

In total, our primary evaluation includes 34 unique attacks. Additionally, we conduct a separate experiment to test watermark robustness in re-embedding scenario. We reimplement each of the 26 post-hoc watermarking methods in our benchmark as attacks and test them across all 32 available methods. 

\textbf{Datasets}. We employ subsets of \textbf{COCO}~\cite{lin2015microsoftcococommonobjects} and \textbf{DiffusionDB} \cite{wangDiffusionDBLargescalePrompt2022} for our evaluations~--- two common datasets in the field of watermarking. To balance computational strain and dataset diversity, we sample 1k source images from each of the two sets for robustness evaluations. Considering the number of watermarking methods and attacks employed, this results in more than 2.8M processed images and $>$8k GPU-hours across our experiments. To evaluate cross-dataset performance, we further employ smaller samples of 400 images from 3 additional datasets: NIPS~\cite{nipsDataset}, KonIQ~\cite{koniq10k}, and DIV2K~\cite{div2k}, with resolutions ranging from 299x299 to 2040x1300.

\begin{figure}[t!]
    \centerline{\includegraphics[width=0.414\textwidth]{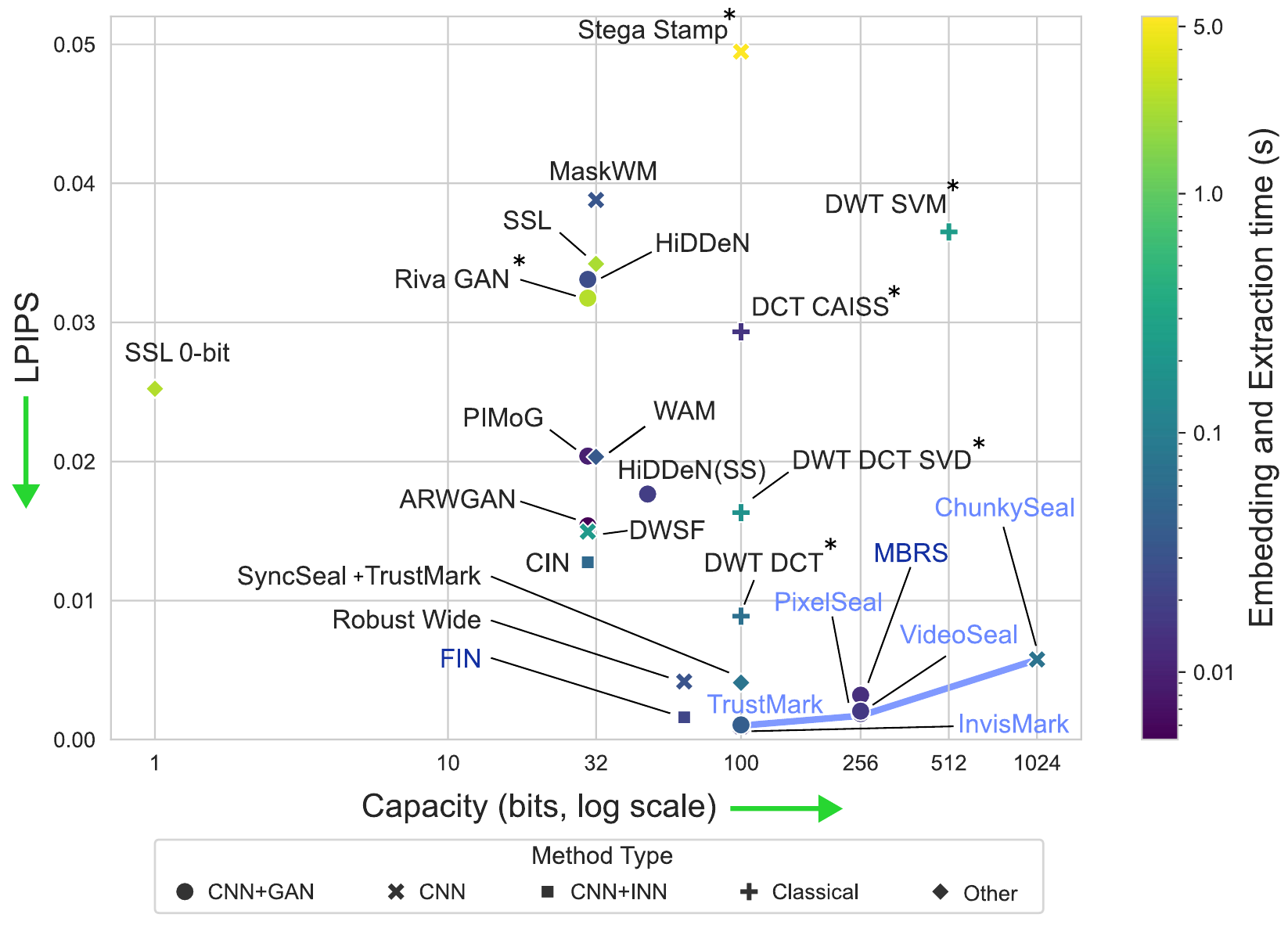}}
    \caption{Capacity-quality tradeoff for all tested post-hoc watermarking methods on clear images from MS-COCO. Color of the point represents execution speed, and marker determines the watermark type. Blue line indicates the Pareto-optimal front. 
    * indicates algorithms that only use CPU.}
    \Description{Scatter plot of embedding capacity in bits on a logarithmic x-axis from 1 to 1024 against LPIPS distortion on the y-axis from 0 to 0.05, where lower is better, for all tested post-hoc watermarking methods on MS-COCO. Marker shape encodes the architecture family and marker colour encodes combined embedding and extraction time. StegaStamp sits alone at the top with by far the highest distortion. Earlier methods such as SSL, HiDDeN, DWT SVM and DWT DCT occupy a low-capacity region with visible distortion, while recent methods including PixelSeal, VideoSeal, MBRS, InvisMark and ChunkySeal form a Pareto front along the bottom right, combining capacities of 100 to 1024 bits with LPIPS below 0.01. Colour also separates the two computational regimes, with the classical CPU-only methods slower than the GPU-accelerated learned ones.}
\label{fig:clear_performance}
\end{figure}

\textbf{Evaluation metrics}. To assess image watermarking methods, we employ two primary classes of metrics. The first assesses watermark robustness, i.e., how accurately the extraction algorithm identifies the presence of a watermark after attacks.

For zero-bit watermarking, where the goal is to detect the presence or absence of a watermark, robustness is assessed with binary classification metrics, such as the \textbf{p-value}, which indicates the confidence in a detection.
In practical applications where false positives are highly undesirable, the metric of choice becomes the True Positive Rate at a specific, low False Positive Rate (\textbf{TPR@x\%FPR}).

In multi-bit watermarking, where a binary payload is embedded, robustness is typically assessed by comparing the extracted message against the original.
The primary metric is the Bit Error Rate (\textbf{BER}), which calculates the proportion of incorrectly extracted bits.
A stricter metric, the Word Error Rate (\textbf{WER}), considers the entire message lost if even a single bit is decoded wrong. 
To compare multi-bit and zero-bit methods fairly, the multi-bit case is commonly reduced to a zero-bit one by thresholding the number of matching bits, enabling unified \textbf{TPR@x\%FPR} reporting. To avoid extensive negative trials, we compute TPR@x\%FPR \textit{theoretically}: the detection threshold for a target FPR follows analytically from the message length $k$ under the null model that non-watermarked bits are i.i.d. Bernoulli(0.5). Further details are provided in Appendix Sec. C.1. 

\begin{figure*}[tbp]
    \centerline{\includegraphics[width=0.99\textwidth]{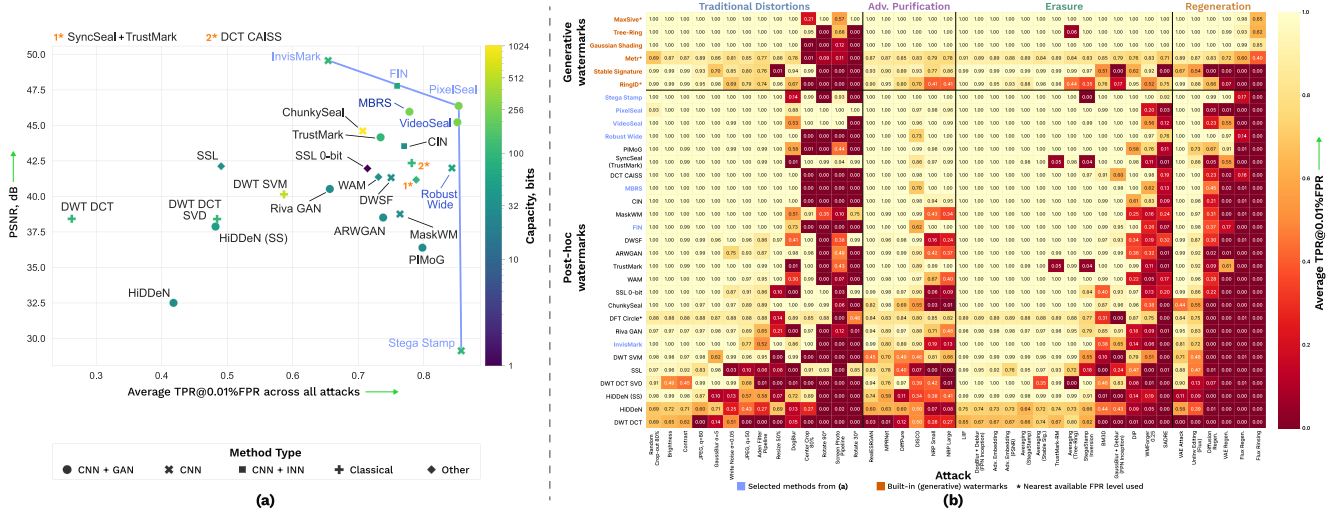}}
    \caption{(a) Robustness-distortion tradeoff for different post-hoc watermarking methods averaged across all tested attacks on DiffusionDB. Robustness is measured with TPR@0.01\%FPR ($\uparrow$); (b) Per-attack robustness breakdown for all tested watermarks.}
    \Description{Two panels. Panel (a) is a scatter plot of average true positive rate at 0.01 percent false positive rate across all attacks on the x-axis against PSNR in decibels on the y-axis, for post-hoc watermarking methods on DiffusionDB, with marker colour encoding embedding capacity. Classical and early methods including DWT DCT, DWT SVM and HiDDeN cluster on the left at low robustness; StegaStamp reaches high robustness but at the lowest PSNR of about 30 decibels; and Robust-Wide, PixelSeal, MaskWM and InvisMark occupy the upper-right frontier, combining high imperceptibility with high robustness. Panel (b) is a heatmap whose rows are all tested watermarking methods, split into a generative block and a post-hoc block, and whose columns are the evaluated attack configurations grouped into traditional distortions, adversarial purification, erasure and regeneration. Cell colour gives the true positive rate at 0.01 percent false positive rate from 0 to 1. Most methods retain high values under traditional distortions and purification, whereas the regeneration columns, the non-right-angle rotation column and SADRE are almost uniformly dark, indicating near-complete watermark loss. The generative watermarks, in particular Gaussian Shading and MaXsive, remain bright across the regeneration columns.}
\label{fig:main_res_diffdb}
\end{figure*}

The second class of metrics addresses image quality.
Full-reference metrics (\textbf{PSNR}, \textbf{SSIM}, \textbf{LPIPS}~\cite{zhang2018unreasonable}, \textbf{DreamSim}~\cite{fu2023dreamsim}) allow a comparison between the original image and the watermarked one for post-generation methods and can also be used to assess the changes introduced by attacks.
No-reference metrics (\textbf{CLIP-IQA}\cite{wang2023exploring}, \textbf{Aesthetics}~\cite{schuhmann2022laion5bopenlargescaledataset}) evaluate the quality of the watermarked image independently, without comparing to the original.
Prompt-based metrics (\textbf{BLIP}~\cite{li2022blip}, \textbf{CLIP Score}~\cite{hessel2021clipscore}, \textbf{Image Reward}~\cite{xu2023imagerewardlearningevaluatinghuman}) assess the alignment of the image with the textual description for in-generation watermarking methods. \textbf{FID}~\cite{heusel2017gans} provides an estimate of the similarity between the distributions of two sets of images.

\section{Results}

This section details the evaluation methodology and presents the obtained results together with a comprehensive analysis.

\subsection{Clear Performance: Trade-offs in image watermarking}
\label{sec:clear_performance}

This experiment evaluates the fundamental steganographic properties of modern watermarking systems, with the implicit research question of whether recent advances in learning-based approaches alter the classical trade-off between embedding capacity, perceptual imperceptibility, and computational efficiency. We benchmark 32 classical, deep learning-based, and generative watermarking methods under a unified protocol, focusing on capacity–quality–efficiency relations. Computational cost is included to reflect deployment constraints, with hardware dependency (CPU vs GPU) explicitly accounted for to ensure comparability across implementations.

Figure \ref{fig:clear_performance} presents the comparison for post-hoc methods, showing that earlier techniques such as SSL, HiDDeN, and DWT DCT follow the classical capacity–imperceptibility trade-off, operating in a low-capacity regime where increased payload leads to visible degradation. In contrast, recent approaches, including PixelSeal, ChunkySeal, and InvisMark, shift this frontier, enabling higher embedding capacities while maintaining or improving perceptual quality, effectively relaxing the traditional trade-off structure. The figure also indicates a separation in computational regimes, where modern methods achieve higher throughput but rely on GPU acceleration, whereas earlier approaches, such as DWT-DCT, remain CPU-efficient but limited in scalability. Generative watermarks are evaluated using No-Reference metrics and FID in Appendix Sec.~C.2. 

Overall, the results indicate that the observed decoupling between capacity and imperceptibility introduces secondary constraints in terms of robustness, which increasingly becomes the primary limiting factor. This observation motivates the subsequent evaluation of robustness under attacks in Section \ref{sec:robustness_eval}.

\subsection{Watermark Robustness: Evaluation in adversarial scenarios}
\label{sec:robustness_eval}

We evaluate all watermarking methods against 34 unique attack
\newpage
\noindent configurations (40, including parameterized variants) under a unified threat model, asking whether any method is consistently resilient across heterogeneous perturbations (RQ1) and how different method classes fail under specific attack families (RQ2) --- i.e., whether robustness generalizes or remains partitioned by method class and threat type. Fig.~\ref{fig:main_res_diffdb} summarizes main results for all tested methods with TPR@FPR metric, and more detailed evaluations with other robustness scores are provided in Appendix Sec.~C.3. 

\textbf{Visibility-capacity-robustness tradeoff. }
Figure~\ref{fig:main_res_diffdb}a demonstrates a structured trade-off between visibility, capacity, and robustness, where modern learned methods dominate the Pareto frontier, replacing classical frequency-domain approaches.  Early methods such as DWT-DCT, DWT-SVM, and HiDDeN occupy the lower-left part of the plot, reflecting their limited robustness even within larger perturbation budgets. In contrast, recent models such as Robust-Wide, PixelSeal, MaskWM, and InvisMark achieve a much better balance between stealthiness and robustness. Among post-hoc methods, StegaStamp is the strongest overall in terms of robustness, while InvisMark stands out as the most imperceptible method that remains on the visibility-robustness frontier. StegaStamp achieves high robustness but at a significantly higher perceptual cost, indicating a less efficient trade-off regime. Taken together, these results suggest that stronger training-time augmentation, and explicit robustness-oriented objectives employed within these methods can substantially improve the quality--robustness balance.

\begin{figure*}[htbp]
    \centerline{\includegraphics[width=0.89\textwidth]{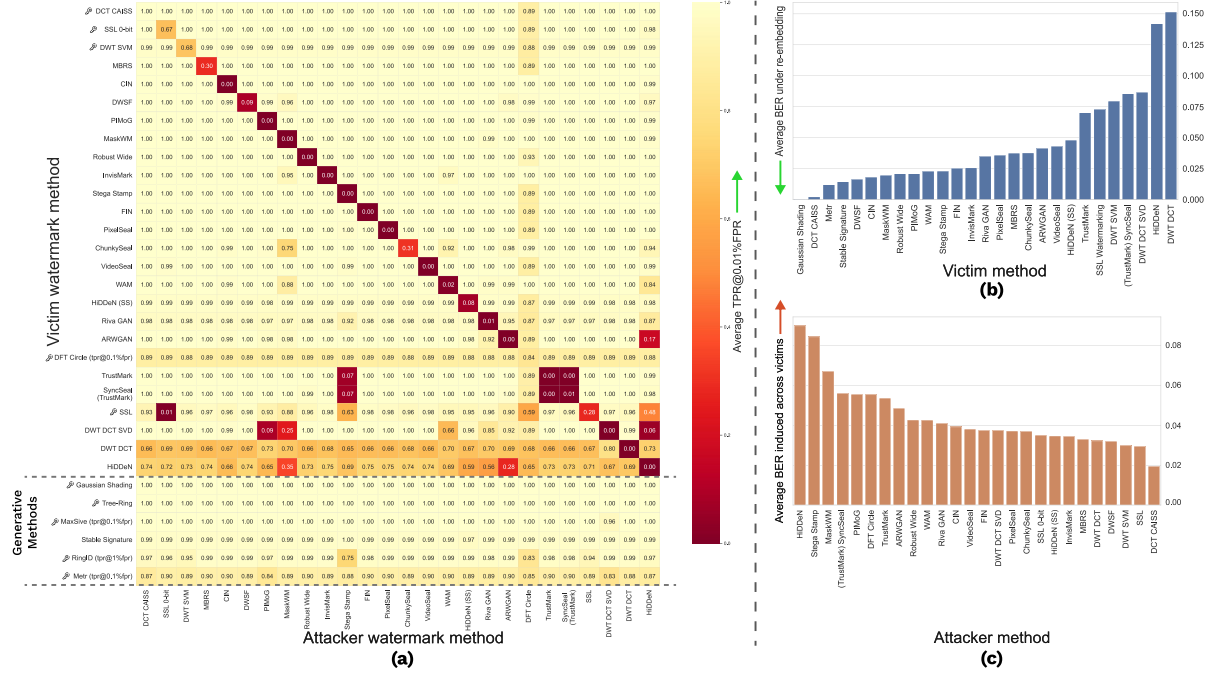}}
    \caption{\textbf{Watermark robustness under re-embedding attacks}. (a) Average TPR@0.01\%FPR ($\uparrow$) measured for each pair of victim and attacker watermarks. (b) Bit Error Rate ($\downarrow$) of multi-bit watermarks under re-embedding attacks, averaged across all attacker watermarks. (c) Watermark efficiency as an attacker is evaluated as the average BER induced across watermarks.}
    \Description{Three panels on re-embedding attacks. Panel (a) is a matrix heatmap whose rows are victim watermarks, with post-hoc methods above and generative methods in a separate block below, and whose columns are attacker watermarks; cell colour gives the average true positive rate at 0.01 percent false positive rate. Most cells are bright, showing broad resistance to external re-embedding, but the diagonal is dark for nearly every post-hoc method, showing that re-applying the same method erases the original mark; key-based methods and DCT CAISS are the exceptions, and the generative rows stay uniformly bright. Panel (b) is a bar chart of the average bit error rate induced under re-embedding for each victim method, sorted in increasing order from Gaussian Shading near zero to DWT DCT at roughly 0.15. Panel (c) is a bar chart of the average bit error rate each attacker method induces across all victims, sorted in decreasing order and led by HiDDeN, StegaStamp, PIMoG and MaskWM, the methods with the lowest PSNR.}
\label{fig:reembed_results}
\end{figure*}
Capacity is observed to improve robustness up to a point, likely due to redundancy effects in decoding, as larger number of bit-swaps is required to reach the same FPR threshold. However, this relationship is non-monotonic, as demonstrated by ChunkySeal, where increasing the payload to 1024 bits does not yield superior robustness compared to more moderate-capacity designs such as PixelSeal or VideoSeal. Under a fixed imperceptibility budget, larger payloads can become harder to protect at the bit level, so robustness depends not only on architectural family but also on how efficiently the available embedding capacity is used.

\textbf{Performance across different threat models. }
Our results highlight a clear separation between traditional distortions and stronger semantic attacks. Standard image manipulations such as moderate JPEG compression, blur, brightness changes, and small crops are tolerated by many modern methods, especially those trained with broad distortion pipelines. In contrast, the most damaging attacks are those that either break spatial alignment or substantially rewrite the image content. Non-right-angle rotations are a particularly revealing failure mode: many otherwise robust watermarks collapse under rotation, showing that geometric synchronization remains a major bottleneck. Methods such as DWSF, MaskWM and DFT-Circle mitigate this limitation via explicit synchronization mechanisms, whereas PixelSeal achieves partial robustness through adversarial training rather than structural alignment modules. The SyncSeal method~\cite{fernandez2025geometric} introduces an external synchronization mechanism that can be applied independently of the underlying watermarking method.

The most severe degradation is observed under regeneration attacks (Fig.~\ref{fig:main_res_diffdb}b). Post-hoc methods are especially vulnerable to attacks that reconstruct or re-synthesize the image, and the FLUX-based regeneration and rinsing attacks are among the most destructive in the benchmark. In these cases, the watermark signal is largely replaced by newly generated content, which leaves very little trace of the original embedding. In contrast, generative watermarks, especially Gaussian Shading and MaXsive, maintain significantly higher robustness under such conditions, indicating a fundamental advantage when the threat model includes image rewriting or synthesis.

\textbf{Cross-dataset evaluation}. We additionally conduct a cross-resolution evaluation on five heterogeneous datasets (Appendix Sec.~C.4
). Overall robustness rankings remain largely consistent across all tested sets. Notably, robustness scores are generally slightly higher at larger resolutions, since many watermarking algorithms operate at fixed low resolutions (typically 224--512 px) and downscale images internally, thereby smoothing out finer adversarial perturbations.

Overall, the main robustness results suggest three practical conclusions, which directly address RQ1 and RQ2. First, the most competitive methods are not the oldest or the highest-capacity ones, but the ones that combine learned embeddings with explicit robustness-oriented training, indicating that consistent resilience across perturbation types is achievable only under specific design choices rather than being a general property of any method class. Second, geometric alignment remains a crucial weakness, and rotation-aware design is still a meaningful architectural advantage, which reflects a systematic and a recurring failure mode under geometric transforms across most method families. Third, post-hoc methods exhibit a notable robustness gap compared to generative approaches, and the choice between the two classes should be guided by the expected threat model and practical use case: post-hoc methods are suitable when the main risk is conventional distortion or non-destructive image processing, whereas generative watermarks are applicable in a limited set of scenarios and are preferable when the adversary can perform strong image reconstruction or regeneration.

\subsection{Like cures Like: Evaluating watermark robustness under re-embedding attacks}
\label{sec:reembed_attacks}

In this experiment, we address the RQ3 and focus on the behavior of watermarks under re-embedding attacks, i.e., when additional watermarks are superimposed onto an existing one. This scenario is both practically relevant and methodologically important, as it reflects realistic conditions in which a third party might re-embed an image originally watermarked by another entity, thereby overwriting it with their own identifier. In such cases, the original watermark must remain readable, preserving its integrity despite the attack. 
The study is conducted on DiffusionDB, with all watermarks evaluated against one another. Post-hoc watermarks with random bit sequences are used as attacks, while built-in watermarks cannot serve as attackers and are therefore tested only as victims. Figure~\ref{fig:reembed_results}a reports average TPR@0.01\%FPR scores across victim-attacker pairs.

\textbf{Post-hoc methods over other post-hoc methods.}
The experimental results show that watermarks are relatively resistant to external post-hoc attacks. The largest increases in BER are observed among methods that already exhibit low robustness to attacks: DWT DCT, HiDDeN, and DWT DCT SVD (see Figure~\ref{fig:reembed_results}b).
Conversely, the methods that induce the most significant BER degradation when acting as attackers tend to be those with the lowest PSNR: HiDDeN, StegaStamp, PIMoG, and MaskWM. The victim-attacker matrix reveals notable asymmetry: "stronger" methods often erase "weaker" ones even when image quality is comparable (Fig. \ref{fig:reembed_results}a). For instance, GAN-based PIMoG invalidates classical DWT-DCT-SVD, while the reverse has minimal impact. Surprisingly, DWT-DCT-SVD resists the more disruptive StegaStamp attacker, suggesting these asymmetries stem from structural frequency-domain differences (Appendix Fig.~19
) rather than simple degradation levels.

\textbf{Resilience to self re-embedding}. The results reveal that most watermarks exhibit limited resilience to self-re-embedding. A second application of the same method effectively removes the original watermark. However, methods that rely on a key (marked with the key icon in Figure~\ref{fig:reembed_results}a) can survive re-embedding provided a different key is used. Notably, only the DCT CAISS method demonstrates strong resistance to self-overlays, an observation consistent with experiments originally reported in the Spread Spectrum Watermarking literature~\cite{cox1997secure}. In contrast, the majority of neural post-hoc methods are keyless and remain vulnerable to self-erasure.

\textbf{The behavior of built-in watermarks under post-hoc overlays.} Built-in watermarks prove to be practically immune to post-hoc overlays. In particular, Stable Signature, obtained by fine-tuning a Stable Diffusion model on the decoder of Hidden (SS), remains robust against attacks from the same method. Despite sharing a decoder with Hidden (SS), Stable Signature effectively withstands re-embedding attacks launched with the Hidden (SS) encoder. 
The only exception among generative methods is METR, which shows a small but consistent drop in TPR under any re-embedding. As illustrated in Fig.~\ref{fig:main_res_diffdb} (b), METR exhibits limited robustness in general. This vulnerability may arise from its low capacity of only 10 bits, where even a single bit flip can significantly degrade the TPR.

\textbf{Potential benefits of combining multiple watermarking methods.} Beyond robustness to re-embedding attacks, the findings also suggest that multiple watermarking methods could potentially be combined within a single image. Such integration could enable an increased embedding capacity or the implementation of dual schemes, where different watermarks serve complementary purposes, for instance, one for ownership verification and another for tracking or authentication. However, combining methods would result in increased perturbation budget, potentially compromising the imperceptibility requirements. Therefore, a more detailed evaluation might be necessary for practical deployment.

\subsection{Adversarial Strength: Evaluating attacks from the attacker’s perspective}
\begin{figure}[t]
    \centerline{\includegraphics[width=0.49\textwidth]{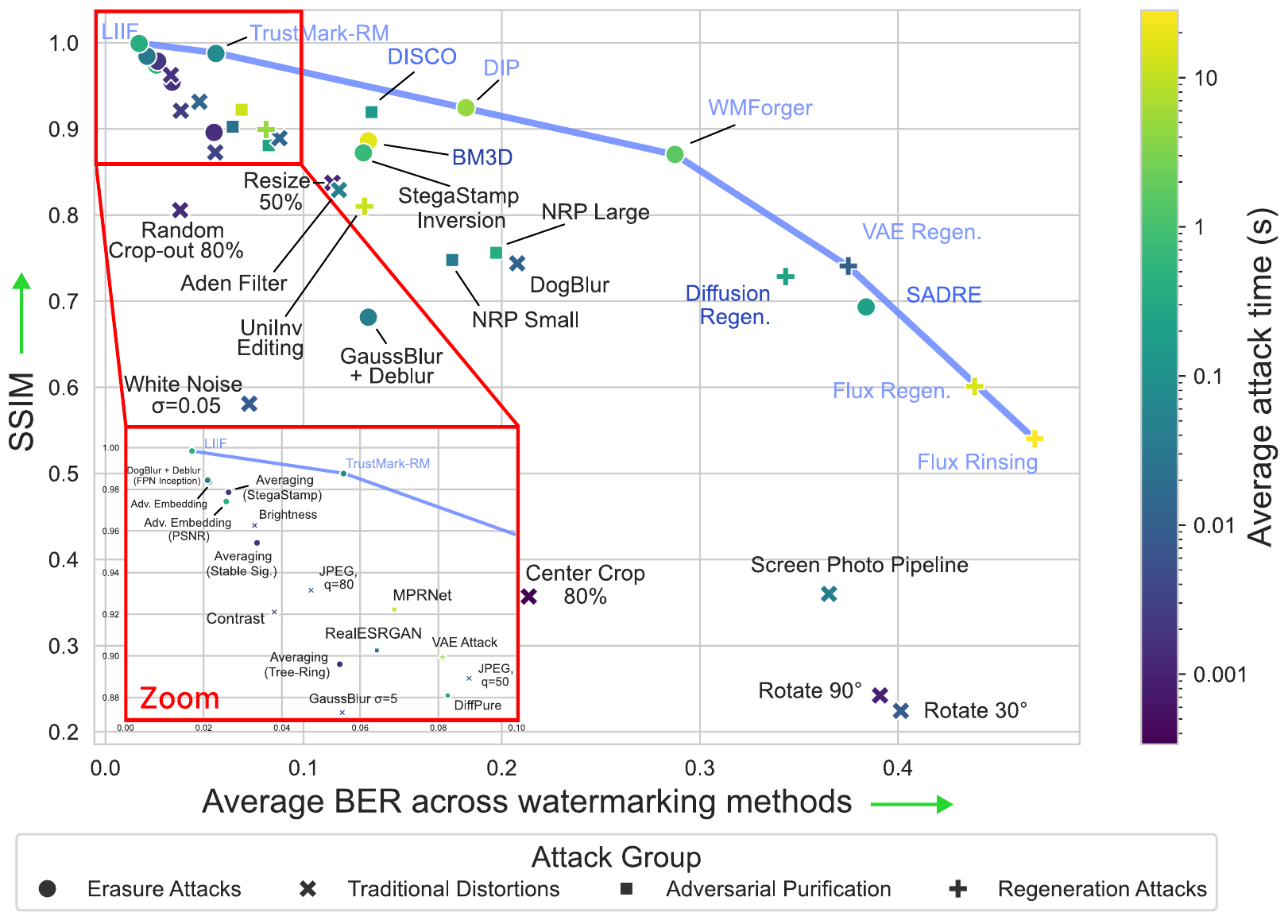}}
    \caption{Efficiency-distortion tradeoff for different attacks averaged across all watermarks on DiffusionDB dataset. Attack efficiency is measured with BER. Attacks with colored names lie close to the efficiency-visibility Pareto front.}
    \Description{Scatter plot of attack effectiveness against image fidelity for all evaluated attacks on DiffusionDB, with average bit error rate across watermarking methods on the x-axis from 0 to about 0.45 and SSIM on the y-axis from 0.2 to 1.0. Marker shape encodes the attack group and colour encodes average runtime; an inset zooms into the crowded low-BER region. Dedicated erasure attacks including WMForger, TrustMark-RM, DISCO and DIP lie along the upper-right Pareto front, combining high bit error rates with minimal image change. Regeneration attacks such as Flux Regeneration, Flux Rinsing and SADRE reach the highest bit error rates but the lowest SSIM values. Geometric transforms, particularly 30 and 90 degree rotation and centre cropping, also combine high bit error rates with low SSIM.}
\label{fig:attack_avg_performance}
\end{figure}

This experiment addresses the second part of RQ2, shifting focus to the attacker’s perspective and analyzing which transformations are most effective at disrupting watermarks. To this end, we aggregate BER induced by each attack across all multi-bit watermarking methods, providing a coarse but informative method-agnostic ranking of attack strength (Fig.~\ref{fig:attack_avg_performance}). 
This analysis is intended to highlight the most critical attack vectors, thereby informing the design of more robust future watermarking systems.

A clear \textbf{Strength/Perceptual Similarity tradeoff} can be observed among attackers (Fig.~\ref{fig:attack_avg_performance}).
Notably, methods explicitly designed for watermark removal consistently dominate this frontier, achieving strong disruption under relatively small perceptual degradation. In particular, WMForger, TrustMark-RM and SADRE are among the best in efficiency, maintaining high attack success while preserving reasonable image fidelity.

Among \textbf{Adversarial Purification methods}, performance varies substantially. DiffPure shows relatively weak effectiveness, suggesting that removing adversarial perturbations does not directly translate to removing structured watermark signals. In contrast, approaches based on implicit neural representations, such as DISCO and DIP, are significantly more effective, aligning with prior observations~\cite{liangBaselineMethodRemoving2025} that such models can implicitly overwrite high-frequency embedded signals. Classical restoration models (e.g., RealESRGAN, MPRNet) also provide some erasure capability, but remain suboptimal compared to dedicated attacks.

\textbf{Traditional distortions} exhibit a different regime: while many common perturbations (e.g., mild compression or noise) are relatively weak, geometric transformations such as rotation and cropping rank among the most destructive, although their pixel-level image quality cannot be well captured with Full-Reference metrics such as SSIM. To better account for perceptual and semantic changes, we additionally evaluate attack quality using CLIP-IQA (No-Reference technical quality score) and an aesthetic predictor, with detailed comparisons provided in the Appendix Sec.~C.5.

Finally, we analyze attack strength as a function of controllable hyperparameters (Appendix Sec.~C.6
). Among the evaluated methods, WMForger demonstrates the most favorable rate--distortion behavior, consistently outperforming alternatives across a wide range of settings. Interestingly, even simple baselines such as JPEG compression retain competitive effectiveness at lower quality factors, highlighting that even traditional distortions can still pose a meaningful threat under strong compression regimes.

\section{Conclusion}
This work presents WARP, a reproducible benchmark for invisible image watermarking that evaluates 32 methods under a diverse set of 34 attack scenarios to expose the trade-offs between capacity, imperceptibility, and robustness. 
No method is uniformly robust~(RQ1): learned and generative approaches improve over baselines but remain vulnerable to geometric, purification, and regeneration attacks, indicating that robustness reflects threat-model alignment rather than intrinsic superiority. We identify structured failure modes~(RQ2), including geometric misalignment, asymmetric re-embedding, and regeneration-based removal. Re-embedding frequently overwrites post-hoc watermarks, while generative and key-based methods resist it~(RQ3), implying that persistence depends on the embedding mechanism. 
In practice, our results favor post-hoc methods with explicit synchronization when the expected threat is conventional processing, and in-generation watermarks wherever the pipeline allows them, as these are the only ones that reliably survive regeneration.

\newpage

\begin{acks}
This work was supported by a grant provided by the Ministry of Economic Development of the Russian Federation in accordance with the subsidy agreement (agreement identifier 000000C313925P4G0002) and the agreement with the Ivannikov Institute for System Programming of the Russian Academy of Sciences dated June 20, 2025, No. 139-15-2025-011. The research was carried out using the MSU-270 supercomputer of Lomonosov Moscow State University and ISP RAS computing cluster.
\end{acks}

\bibliographystyle{ACM-Reference-Format}
\balance
\bibliography{reference}

\newpage
\clearpage
\appendix

\section{Contents}

\begin{itemize}
    \item \textbf{Section \ref{sec:details}} describes all watermarking methods and attacks employed in the benchmark, summarized in Tables~2 and~4 respectively, along with the unified evaluation protocol used across all experiments.
    \item \textbf{Section \ref{sec:further_details_extraction}} provides further details on watermark extraction success metrics, including a validation of the theoretical FPR estimation against an empirical reference set, an analysis of the relationship between TPR@x\%FPR and BER, and a heatmap of TPR across methods and attacks at multiple FPR thresholds.
    \item \textbf{Section \ref{sec:gen_eval}} reports image generation quality evaluations for built-in watermarking methods, comparing watermarked and non-watermarked outputs from the same generative model across FID and five per-image quality metrics.
    \item \textbf{Section \ref{sec:full_wm_attack}} presents the full per-method, per-attack robustness breakdown across all 32 watermarking methods and 34 attack configurations, measured by Average BER, WER and TPR@0.01\%FPR.
    \item \textbf{Section \ref{sec:cross_dset}} reports cross-dataset evaluation results and compares watermark robustness across different resolutions ranging from 299x299 to 2040x1300.
    \item \textbf{Section \ref{sec:metrics}} evaluates perceptual and semantic changes introduced by watermarking methods and attacks, using four complementary imperceptibility metrics for watermarks and three quality metrics for attacks.
    \item \textbf{Section \ref{sec:attack_hyp}} analyzes attack strength as a function of controllable hyperparameters for six selected attacks, reporting per-method BER–distortion curves and their cross-method averages.
    \item \textbf{Section \ref{sec:freq_examples}} visualizes the frequency-domain signatures of all post-hoc watermarking methods via averaged FFT spectrum differences.
    \item \textbf{Section \ref{sec:limitations}} discusses the limitations of the current benchmark and outlines directions for future work.
\end{itemize}

\section{Details on employed methods}
\label{sec:details}

Table \ref{tab:watermarks_table} summarizes the invisible watermarking methods utilized in our study. Table \ref{tab:attacks_table} describes the adversarial attacks used to test the robustness of the watermarking algorithms.

We adopt a unified evaluation protocol to ensure a fair and reproducible comparison across all watermarking methods. During evaluation, a random bit sequence is generated independently for each input image and used as the watermark payload.

To ensure consistent spatial alignment across methods and support arbitrary image resolutions, we follow the resolution scaling strategy proposed in TrustMark~\cite{bui2023trustmark}. Specifically, each image is first resized to a fixed resolution for watermark embedding, after which the residual between the watermarked and original images is computed and upscaled back to the original resolution, and added to the input image. This approach allows all methods to operate at their native resolution while preserving the structure of the embedded signal.

We evaluate all methods on two datasets: DiffusionDB~\cite{wangDiffusionDBLargescalePrompt2022} and MS-COCO~\cite{lin2015microsoftcococommonobjects}. For each method, we use the official open-source implementations provided by the authors, which are integrated into our framework as submodules. All experiments are conducted using PyTorch, and pretrained weights released by the original authors are used without additional fine-tuning. The exception is the HiDDeN method: in addition to the base model, the Hidden (SS) version provided by the authors of Stable Signature is also used. For classical watermarking methods that do not have publicly available implementations, including DFT Circle, DWT SVM, and DCT CAISS, we provide our own implementations based on the descriptions in the corresponding papers. All preprocessing, post-processing, and evaluation procedures are applied uniformly across all methods to ensure a consistent and fair benchmarking setup.

\begin{table*}[h]
\caption{Watermarking methods and their properties. \textit{Type} denotes whether the watermarking algorithm processes an existing image or is built into the process of image generation. \textit{Local/global} denotes whether watermark affects only a portion of / all pixels (if it is added in pixel space) or coefficients of the domain it is inserted in (e.g. Fourier coefficients or latent space). \textit{Video} denotes whether the algorithm was designed to support video in addition to images.}
\centering
\resizebox{2.1\columnwidth}{!}{
\begin{tabular}{lcccccccccc}
\toprule

Name & Year & Type & Binary message & Bits & NN-based & Architecture & Frequency Domain & Latent Space & Local/Global & Video  \\
\midrule

DWT DCT \cite{al2007combined} & 2007 & post-hoc & yes & 100 bits & no & - & yes (DWT, DCT) & no & local & no \\
DWT DCT SVD \cite{navas2008dwt} & 2008 & post-hoc & yes & 100 bits & no & - & yes (DWT, DCT) & no & local & no \\
DFT Circle \cite{poljicak2011discrete} & 2011 & post-hoc & no & zero-bit & no & - & yes (FFT) & no & local & no \\
DCT CAISS \cite{guzik2015real} & 2011 & post-hoc & yes & 800 bits & no & - & yes (DCT) & no & local & no \\
DWT SVM \cite{islam2020svm} & 2020 & post-hoc & yes & 512 bits & no & Support Vector Machine & yes (DWT) & no & local & no \\
HiDDeN \cite{zhu2018hidden} & 2018 & post-hoc & yes & 30 bits & yes & CNN, GAN & no & no & global & no \\
RivaGAN \cite{zhang2019robust} & 2019 & post-hoc & yes & 30 bits & yes & CNN, GAN, Attention & no & no & global & yes \\
StegaStamp \cite{tancik2020stegastamp} & 2020 & post-hoc & yes & 100 bits & yes & CNN & no & no & global & no \\
MBRS \cite{jia2021mbrs} & 2021 & post-hoc & yes & 30/256 bits & yes & CNN, GAN & no & no & global & no \\
SSL \cite{fernandez2022sslwatermarking} & 2021 & post-hoc & optional & zero-bit / 32 bits & yes & CNN, DINO & no & yes & local & no \\
CIN \cite{ma2022towards} & 2022 & post-hoc & yes & 30 bits & yes & CNN, Invertible Network & no & no & global & no \\
PIMoG \cite{fang2022pimog} & 2022 & post-hoc & yes & 30 bits & yes & CNN, GAN & no & no & global & no \\
ARWGAN \cite{huang2023arwgan} & 2023 & post-hoc & yes & 30 bits & yes & CNN, GAN & no & no & global & no \\
DWSF \cite{guo2023practical} & 2023 & post-hoc & yes & 30 bits & yes & CNN, UNET & no & no & global & no \\
FIN \cite{fang2023flow} & 2023 & post-hoc & yes & 64 bits & yes & Flow-based Model, Invertible Network & no & no & global & no \\
SS HiDDeN \cite{fernandez2023stable} & 2023 & post-hoc & yes & 48 bits & yes & CNN & no & no & global & no \\
TrustMark \cite{bui2023trustmark} & 2023 & post-hoc & yes & 100 bits & yes & CNN, GAN & no & no & global & no \\
InvisMark \cite{xu2025invismark} & 2024 & post-hoc & yes & 100 bits & yes & CNN & no & no & global & no \\
Robust-Wide \cite{hu2024robust} & 2024 & post-hoc & yes & 64 bits & yes & CNN, UNET & no & no & global & no \\
VideoSeal \cite{fernandez2024video} & 2024 & post-hoc & yes & 256 bits & yes & CNN, UNET, ViT, GAN & no & no & global & yes \\
Watermark Anything \cite{sander2024watermark} & 2024 & post-hoc & yes & 32 bits & yes & VAE & no & yes & local & no \\
ChunkySeal \cite{petrov2025we} & 2025 & post-hoc & yes & 1024 bits & yes & CNN, UNET & no & no & global & yes \\
MaskWM \cite{hu2025mask} & 2025 & post-hoc & yes & 32/64/128 bits & yes & CNN, UNET & no & no & both & no \\
PixelSeal \cite{souvcek2025pixel} & 2025 & post-hoc & yes & 256 bits & yes & CNN, UNET, GAN & no & no & global & yes \\
SyncSeal \cite{fernandez2025geometric} & 2025 & post-hoc & method-dependent & method-dependent & yes & CNN, UNET, GAN & no & no & global & no \\
Stable Signature \cite{fernandez2023stable} & 2023 & built-in & yes & 48 bits & yes & Diffusion, VAE & no & no & global & no \\
TreeRing \cite{wen2023tree} & 2023 & built-in & no & zero-bit & yes & Diffusion, VAE & yes (FFT) & yes & local & no \\
Gaussian Shading \cite{yang2024gaussian} & 2024 & built-in & yes & 256 bits & yes & Diffusion, VAE & no & yes & global & no \\
METR \cite{varlamov2024metr} & 2024 & built-in & yes & 10 bits & yes & Diffusion, VAE & yes (FFT) & yes & local & no \\
Ring-ID \cite{ci2024ringid} & 2024 & built-in & no & zero-bit & yes & Diffusion, VAE & yes (FFT) & yes & local & no \\
MaXsive \cite{mao2025maxsive} & 2025 & built-in & no & zero-bit & yes & Diffusion, VAE & yes (FFT) & yes & local & no \\

\bottomrule
\end{tabular}

}
\label{tab:watermarks_table}
\end{table*}

\section{Additional results}
\label{sec:add_results}
\begin{figure}[tbp]
    \centerline{\includegraphics[width=0.99\columnwidth]{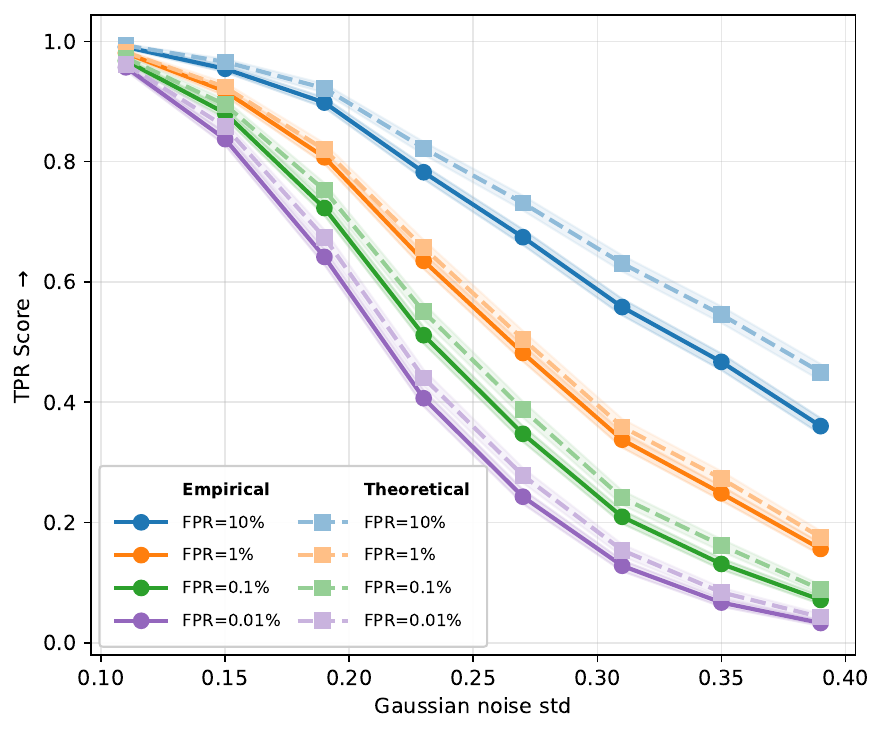}}
    \caption{ Theoretical vs. empirical TPR@FPR for multibit watermarking under Gaussian noise. Solid lines represent empirical TPR estimated using a reference set of N=100,000 non-watermarked images from MS-COCO train split following the WAVES protocol; Dashed lines represent theoretical TPR computed from the binomial model assuming i.i.d. Bernoulli(0.5) bits. Four FPR thresholds are shown: $10\%$, $1\%$, $0.1\%$ and $0.01\%$. Results are reported for TrustMark (k=100 bits) on the MS-COCO validation split.}
    \Description{Line plot of TPR on the y-axis from 0 to 1 against Gaussian noise standard
deviation on the x-axis from 0.10 to 0.40, for TrustMark with a 100-bit payload on the
MS-COCO validation split. Four pairs of curves correspond to false positive rate
thresholds of 10 percent, 1 percent, 0.1 percent and 0.01 percent; solid lines give the
empirical TPR estimated from a reference set of 100,000 non-watermarked images and
dashed lines the theoretical value from the binomial model. All curves decrease
monotonically as noise increases, and within each pair the theoretical curve lies
slightly above the empirical one, with the gap widening at higher noise levels and at
looser thresholds.}
\label{fig:empirical_theretical_tpr_fpr}
\end{figure}

\subsection{Further details on extraction success metrics}
\label{sec:further_details_extraction}

\textbf{Detection evaluation for multi-bit Watermarking.} Following the detection protocol in Sec.~\ref{sec:bench}, we evaluate multi-bit watermarking methods using the True Positive Rate at a given False Positive Rate (TPR@xFPR). For a \(k\)-bit embedded message  \(m\) and an extracted message \(m'\), detection is based on the number of matching bits \(M(m, m')\). In this work, we adopt the \textbf{theoretical} FPR estimation widely used in previous works~\cite{fernandez2023stable, yu2021artificial, kim2024wouaf}. Under the null hypothesis that the image is non-watermarked, the extracted bits are assumed i.i.d. Bernoulli(0.5), giving \(\mathrm{FPR}(\tau) = I_{1/2}(\tau+1, k-\tau)\) for a decision threshold \(\tau\), where $I_x(a,b) $ is the regularized incomplete beta function. This allows computing arbitrarily low FPR values (e.g., \(10^{-6}\)) without requiring massive non-watermarked datasets.

An alternative \textbf{empirical} estimation, used in benchmarks like WAVES~\cite{an2024waves}, constructs a reference set of non-watermarked images. For a test image, the empirical FPR is the fraction of reference extractions whose Hamming distance to \(m'\) is no larger than \(d_H(m', m)\). While this method makes no distributional assumptions, it requires \(N \gg 1/\mathrm{FPR}\) reference images and becomes infeasible for very low FPR targets (e.g. $10^{-6}$ FPR threshold would require at least $10^7$ samples).

To validate the theoretical assumption under WARP conditions, we conduct a direct comparison. We select a representative multibit method (TrustMark with \(k=100\) bits) and collect \(N=100,\!000\) non-watermarked images from the MS-COCO training split~\cite{lin2015microsoftcococommonobjects} to build the empirical reference. \(5,\!000\) images from validation split are watermarked with random Bernoulli(0.5) messages. We then apply Gaussian noise of increasing intensity to simulate degradation, measuring TPR at a fixed target \(\mathrm{FPR}=10^{-3}\). The results in figure~\ref{fig:empirical_theretical_tpr_fpr} show that for $FPR \le 1\%$ the theoretical TPR and the empirical TPR  differ slightly across all noise levels, despite a chi-square test confirming that the extracted bits are not perfectly i.i.d. Bernoulli.

\begin{figure}[htbp]
    \centerline{\includegraphics[width=0.99\columnwidth]{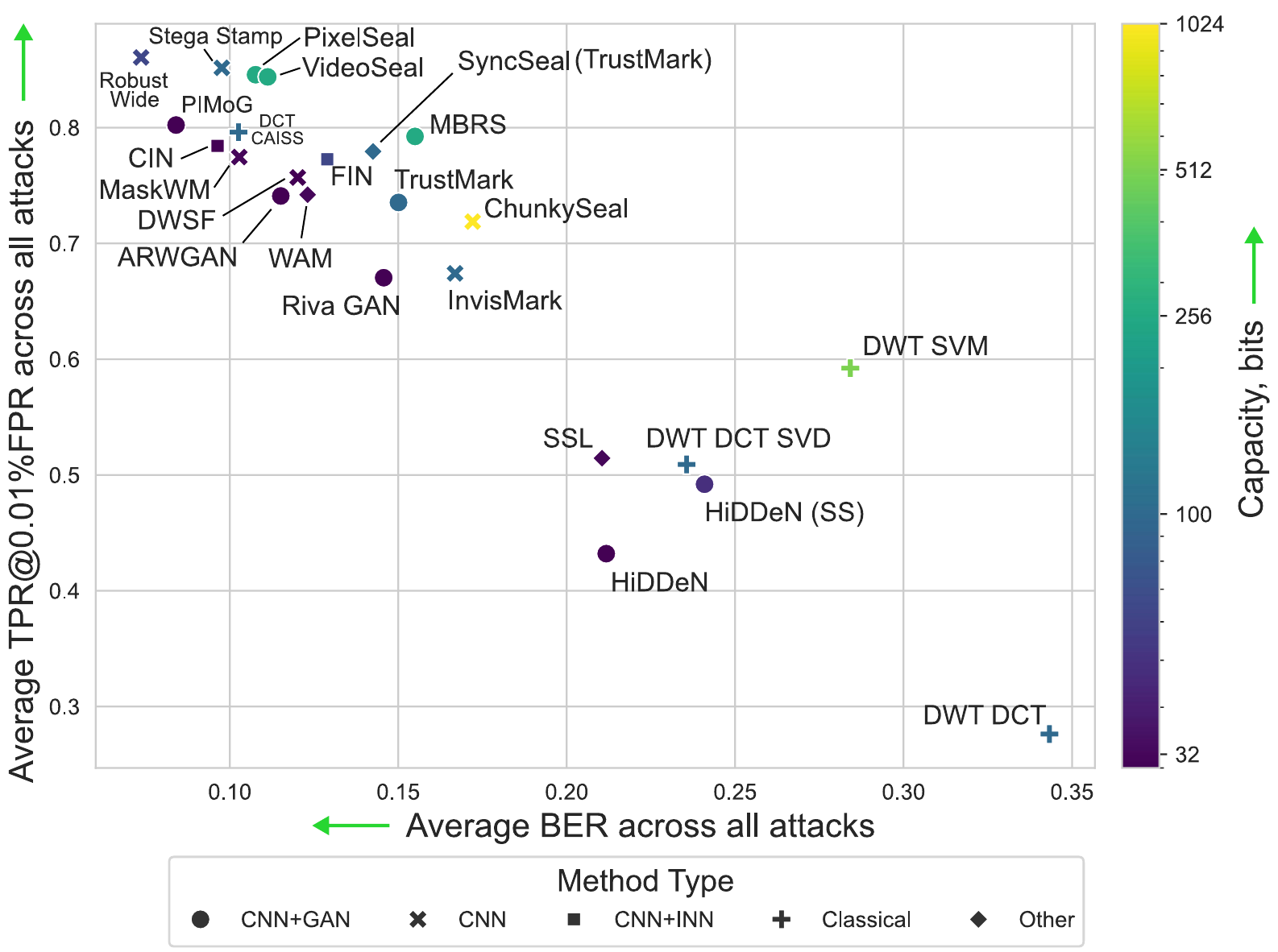}}
    \caption{ TPR@0.01\%FPR vs. Bit Error Rate (BER). Each point corresponds to a multibit watermarking method. The x-axis reports BER (lower is better), the y-axis reports TPR at a fixed FPR of 0.01\% (higher is better). Marker colors denote different payload capacities. Results are aggregated across all evaluated attacks on the DiffusionDB dataset.}
    \Description{Scatter plot with average bit error rate across all attacks on the x-axis
from 0.10 to 0.35, where lower is better, and true positive rate at a fixed 0.01
percent false positive rate on the y-axis, with one point per multi-bit watermarking
method and colour encoding payload capacity. The overall relationship is negative, but
at comparable bit error rates the high-capacity methods, including ChunkySeal with 1024
bits and PixelSeal with 256 bits, achieve substantially higher true positive rates than
30-bit methods such as ARWGAN and Riva GAN. DWT DCT and HiDDeN sit in the lower-right
corner with the worst values on both axes.}
\label{fig:ber_tpr_scatter}
\end{figure}

\textbf{Correlation Between TPR@xFPR and Bit Error Rate (BER).} Figure~\ref{fig:ber_tpr_scatter} shows a clear negative correlation: lower BER corresponds to higher TPR@0.01\%FPR, as expected. However, a more nuanced pattern emerges when considering payload capacity. Methods with larger bit lengths (e.g., ChunkySeal with 1024 bits, PixelSeal with 256 bits) achieve higher TPR at comparable BER than low-capacity methods (e.g., ARWGAN with 30 bits, Riva GAN with 30 bits).

This discrepancy stems from the detection rule in the zero-bit scenario. For a fixed target FPR, the allowable number of mismatched bits grows with bit length $k$, and this growth is nonlinear due to the concentration of measure in the binomial distribution. Larger $k$ permits a higher absolute number of bit errors while still satisfying the same FPR constraint, directly boosting TPR. Thus, TPR@xFPR comparisons across methods with different capacities must account for this effect.

\begin{table*}[ht]
\caption{Image generation quality evaluation for built-in watermarking methods. For each method, images are generated both with and without the watermark using the same underlying generative model. FID is computed against COCO validation images. Val. column represents the absolute value of the corresponding metric, and Rel. $\Delta$ indicates the change relative to the base generative model. See Section \ref{sec:gen_eval} for more details on $\Delta$ scores calculation.}
\centering
\resizebox{2.1\columnwidth}{!}{
\begin{tabular}{l|ccc|cc|cc|cc|cc|cc}
\toprule
{} & \multicolumn{3}{c}{FID} & \multicolumn{2}{c}{CLIP-IQA} & \multicolumn{2}{c}{Aesthetic} & \multicolumn{2}{c}{ImageReward} & \multicolumn{2}{c}{CLIPScore} & \multicolumn{2}{c}{BLIP} \\
{} & Marked & Non-Marked & $\Delta$$\downarrow$ & Val.$\uparrow$ & Rel. $\Delta$, \% $\uparrow$ & Val.$\uparrow$ & Rel. $\Delta$, \% $\uparrow$ & Val.$\uparrow$ & Rel. $\Delta$, \% $\uparrow$ & Val.$\uparrow$ & Rel. $\Delta$, \% $\uparrow$ & Val.$\uparrow$ & Rel. $\Delta$, \% $\uparrow$ \\
Watermark               &        &            &                      &                &                              &                &                              &                &                              &                &                              &                &                              \\
\midrule
Stable Signature &  23.29 &      23.49 &                -0.20 &           0.86 &                        -4.24 &           5.34 &                        -0.45 &           0.45 &                         0.30 &           0.27 &                         0.58 &           0.54 &                         0.25 \\
Gaussian Shading &  25.17 &      25.27 &                -0.09 &           0.92 &                        -0.10 &           5.27 &                         0.05 &           0.42 &                        -0.20 &           0.26 &                        -0.12 &           0.53 &                        -0.22 \\
RingID           &  25.19 &      25.43 &                -0.24 &           0.92 &                        -0.20 &           5.27 &                        -0.16 &           0.42 &                        -0.02 &           0.26 &                        -0.06 &           0.53 &                        -0.23 \\
MaxSive          &  25.41 &      25.27 &                 0.14 &           0.92 &                         0.25 &           5.25 &                        -0.52 &           0.42 &                        -0.09 &           0.27 &                         0.12 &           0.53 &                        -0.04 \\
Metr             &  27.86 &      29.68 &                -1.82 &           0.92 &                        -0.39 &           5.18 &                        -1.44 &           0.38 &                        -2.49 &           0.26 &                        -0.46 &           0.52 &                        -1.76 \\
TreeRing         &  28.81 &      29.75 &                -0.94 &           0.93 &                         0.89 &           5.25 &                         0.24 &           0.49 &                         0.38 &           0.27 &                         0.73 &           0.53 &                         1.02 \\
\bottomrule
\end{tabular}

}
\label{tab:generative_wms}
\end{table*}

\begin{figure*}[htbp]
\centerline{\includegraphics[width=0.67\textwidth]{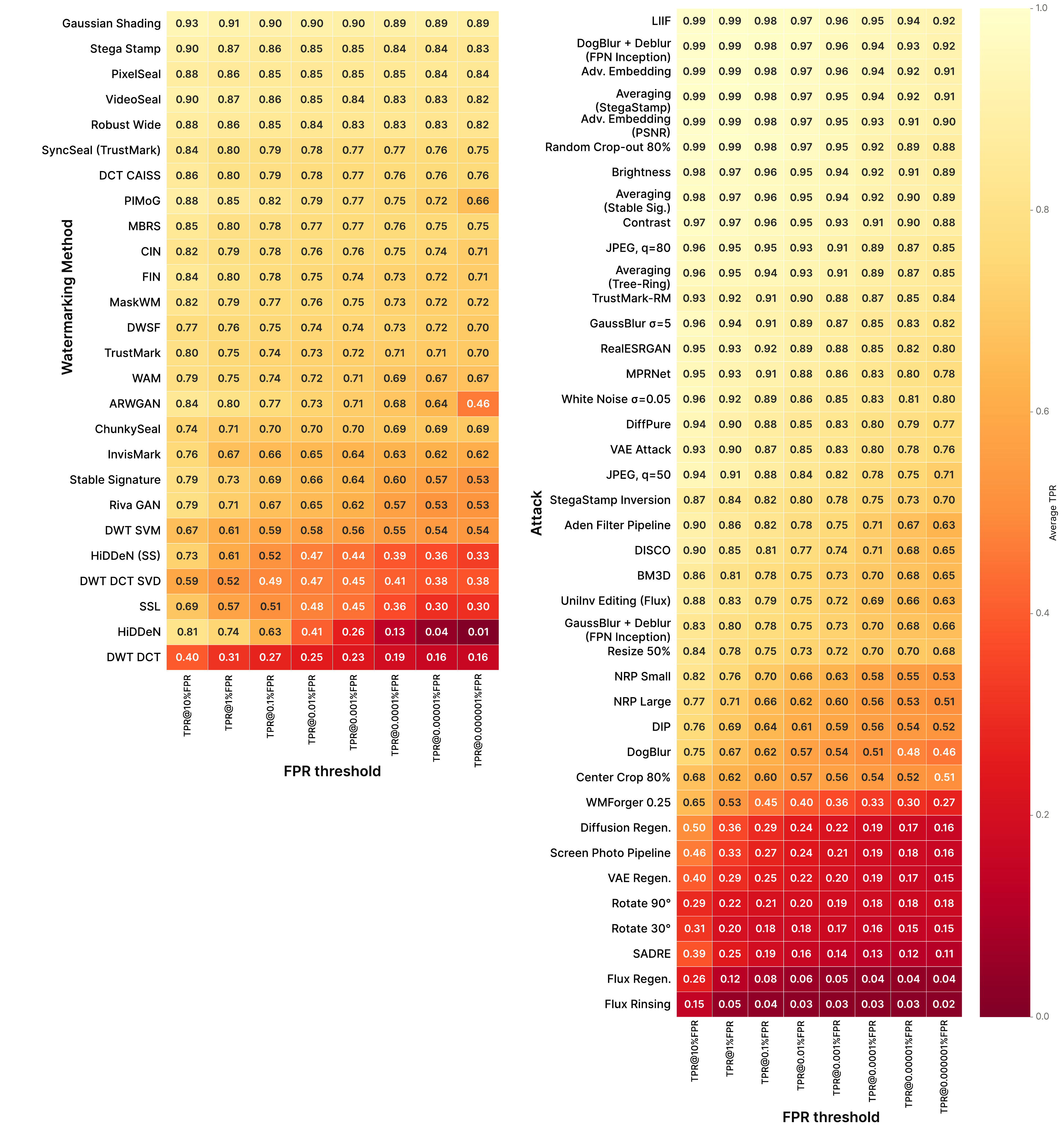}}
    \caption{TPR@xFPR heatmaps for multibit watermarking methods. Two heatmaps show TPR@xFPR across decreasing FPR thresholds from $10^{-1} $ to $10^{-8}$. Left: method-wise heatmap. Rows are multibit watermarking methods, sorted by decreasing average TPR across all attacks and FPR levels. Columns are FPR thresholds. Each cell shows TPR at the corresponding FPR. Right: attack-wise heatmap. Rows are attack types, sorted by decreasing average TPR across all methods and FPR levels. Columns represent FPR thresholds. Each cell shows TPR averaged over all multibit watermarking methods for that attack at the given FPR. Results are reported on the DiffusionDB dataset.}
    \Description{Two heatmaps sharing a column axis of eight false positive rate thresholds
decreasing from 10 percent to 0.000001 percent. In the left heatmap, rows are multi-bit
watermarking methods sorted by decreasing average true positive rate, with Gaussian
Shading, StegaStamp, PixelSeal and VideoSeal at the top and SSL, HiDDeN and DWT DCT at
the bottom. Values fall as the threshold tightens, sharply for low-capacity methods such
as HiDDeN, which drops from about 0.8 to near 0, and only mildly for high-capacity
methods such as ChunkySeal and PixelSeal. In the right heatmap, rows are attacks sorted
by decreasing average true positive rate across methods; LIIF, deblurring and averaging
attacks at the top leave the true positive rate close to 1, whereas rotation, SADRE,
Flux Regeneration and Flux Rinsing at the bottom drive it close to 0 at every threshold.}
\label{fig:heatmap_different_tprs}
\end{figure*}

\textbf{TPR at different FPR.} Figure~\ref{fig:heatmap_different_tprs} presents two complementary heatmaps for multi-bit watermarking methods under the full suite of 34 attacks, with FPR thresholds decreasing from $10^{-1} $ down to $10^{-8}$ .
As FPR decreases, TPR drops for all methods (left heatmap), an expected consequence of tighter statistical thresholds. However, the severity of this drop varies dramatically across methods. HiDDeN (30 bits) exemplifies the most vulnerable class: its TPR falls from 0.82 at $FPR=10^{-1}$ to just 0.01 at $FPR=10^{-8}$, rendering it practically unusable for applications requiring very low false positive rates. ARWGAN exhibits a similar but somewhat less pronounced decline. This fragility is directly linked to payload capacity: low-bit methods have less slack in the binomial detection rule, meaning that even a few bit errors push the matching score below the stringent threshold required for low FPR. High-capacity multibit methods (e.g., ChunkySeal, PixelSeal) maintain substantially higher TPR across the entire FPR range, demonstrating that larger payloads provide inherent statistical advantages for detection.

Averaging TPR across all multi-bit methods per attack (right heatmap) exposes clear differences in attack severity. Regeneration-based attacks (Flux Regeneration, Flux Rinsing, SADRE) and geometric distortions (rotation, cropping) consistently yield the lowest TPR at every FPR threshold, confirming them as the most challenging threat models for current watermarking techniques. In contrast, traditional distortions such as mild JPEG compression and brightness adjustment have the weakest average impact, suggesting that many modern methods have successfully learned robustness against these common perturbations.

\subsection{Image quality evaluations for generative watermarks}
\label{sec:gen_eval}

Table \ref{tab:generative_wms} reports generation quality metrics for all six built-in watermarking methods evaluated on 5k images generated from COCO prompts. Since each method is implemented on top of a different base generative model with different default hyperparameters, absolute metric values are not directly comparable across methods. We therefore focus on the difference between watermarked and non-watermarked outputs from the same underlying model. For FID, we report the raw delta $\Delta = FID(\text{marked}) - FID(\text{non-marked})$, where negative values indicate that the watermarked image distribution is actually closer to the COCO validation set than the base model's output. For all other metrics -- CLIP-IQA, Aesthetic, ImageReward, CLIPScore, and BLIP -- we report absolute values alongside a normalized relative delta (Rel. $\Delta$ in Table \ref{tab:generative_wms}), computed as the mean per-image difference between marked and non-marked outputs divided by the range of non-marked scores, expressed as a percentage: 

$$\text{Rel.}\ \Delta = \frac{\displaystyle\frac{1}{N}\sum_{i=1}^{N}\bigl(s_i^{\text{marked}} - s_i^{\text{non-marked}}\bigr)}{\text{Range}(s^{\text{non-marked}})} \times 100\%$$

where $s_i$ denotes the metric value for image $i$, $N$ is the number of generated images, and $\text{Range}(s)=\max_{i=1,..,N}(s_i)-\min_{i=1,..,N}(s_i)$ represents the metric range on non-marked images. Under this normalization, positive values indicate that inserting the watermark marginally improves the metric relative to the base model. 

The results reveal that built-in watermarks introduce negligible quality degradation and are in most cases practically indistinguishable from their non-watermarked counterparts. Across all per-image metrics, relative deltas remain within a few percent in either direction, with no consistent directional trend -- some methods improve slightly, others degrade slightly, and the differences are well within the noise level expected from stochastic generation. In terms of FID, five of the six methods actually improve upon the base model, in some cases substantially: METR achieves a delta of -1.82 and TreeRing of -0.94, suggesting that their conditioning or noise modification incidentally steers generation toward the target distribution. The only exception is MaXsive, which shows a marginal FID increase of +0.14, remaining negligible in absolute terms. The CLIP-IQA and Aesthetic deltas for Stable Signature are slightly more negative than other methods (-4.24\% and -0.45\% respectively), which may reflect the partial fine-tuning of the decoder that this method relies on, introducing subtle distributional shifts not captured by FID alone. Overall, these results confirm that modern in-generation watermarking methods impose no meaningful quality penalty on the images they protect. Images generated with different built-in watermarks and corresponding non-marked generations are demonstrated in Figure \ref{fig:examples_gen_wms}. 

\subsection{Detailed watermark robustness results}
\label{sec:full_wm_attack}

\begin{figure*}[htbp]
    \centerline{\includegraphics[width=0.85\textwidth]{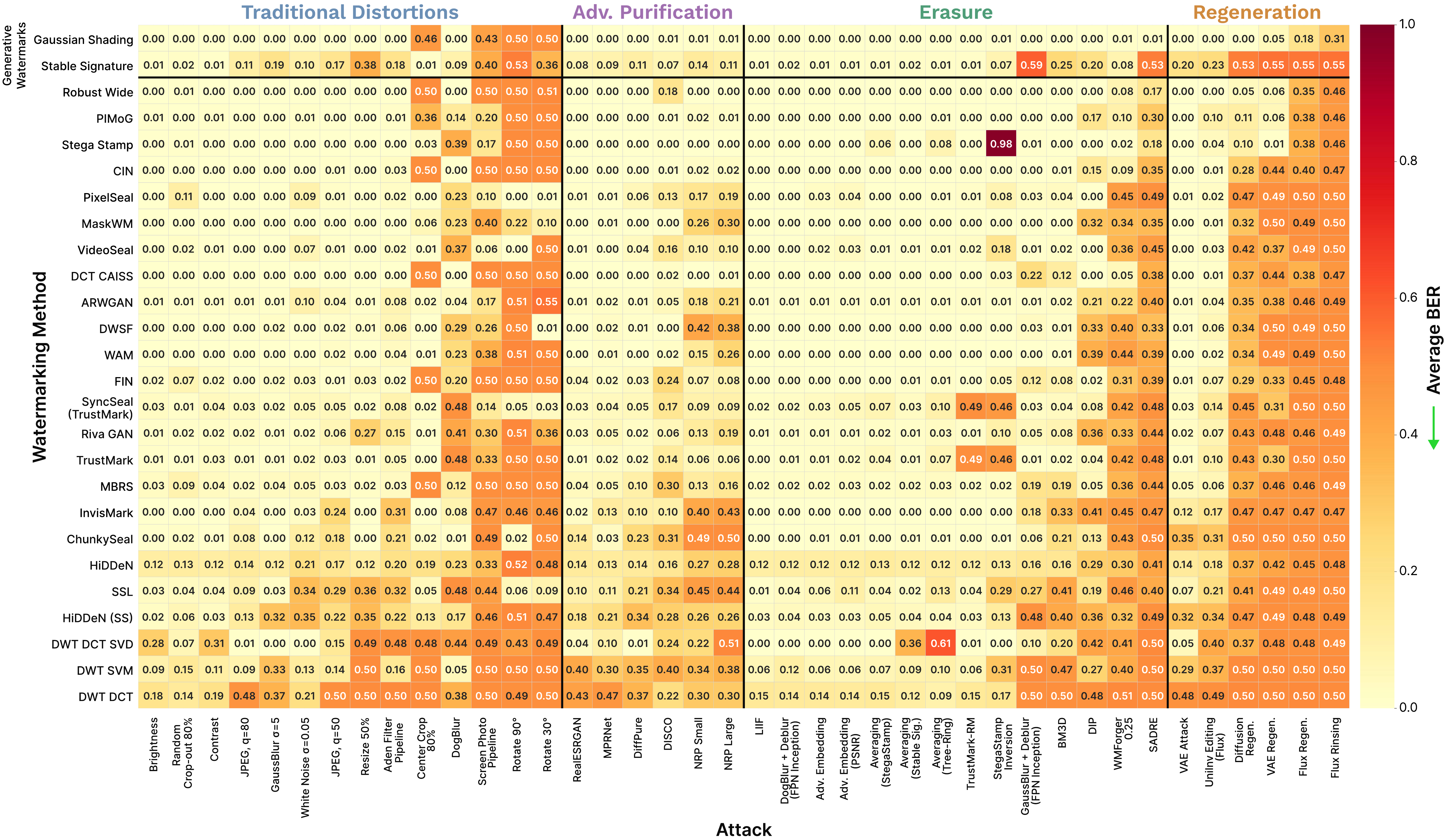}}
    \caption{Full robustness breakdown measured by Average BER ($\downarrow$) for all evaluated watermarking methods across all 40 attack configurations on the DiffusionDB dataset. Each cell reports the mean BER over 1k images; values close to 0.5 indicate complete watermark erasure. Watermarks are sorted by average BER, and attacks are arranged by their average adversarial effectiveness.}
    \Description{Heatmap of average bit error rate on the DiffusionDB dataset. Rows are all
evaluated watermarking methods, with generative watermarks in a separate block at the
top and post-hoc methods sorted by increasing average bit error rate; columns are the
attack configurations, grouped into traditional distortions, adversarial purification,
erasure and regeneration and ordered by increasing adversarial effectiveness. Cells
near 0 are pale and cells near 0.5, corresponding to complete watermark erasure, are
dark. Most cells in the traditional distortion and purification groups are close to 0,
while the regeneration columns and the rotation, SADRE and WMForger columns approach
0.5 for nearly all post-hoc methods. The DWT DCT, DWT SVM and HiDDeN rows are the
darkest overall, and the Gaussian Shading and Robust-Wide rows the palest.}
\label{fig:heatmap_ber}
\end{figure*}

\begin{figure*}[htbp]
    \centerline{\includegraphics[width=0.85\textwidth]{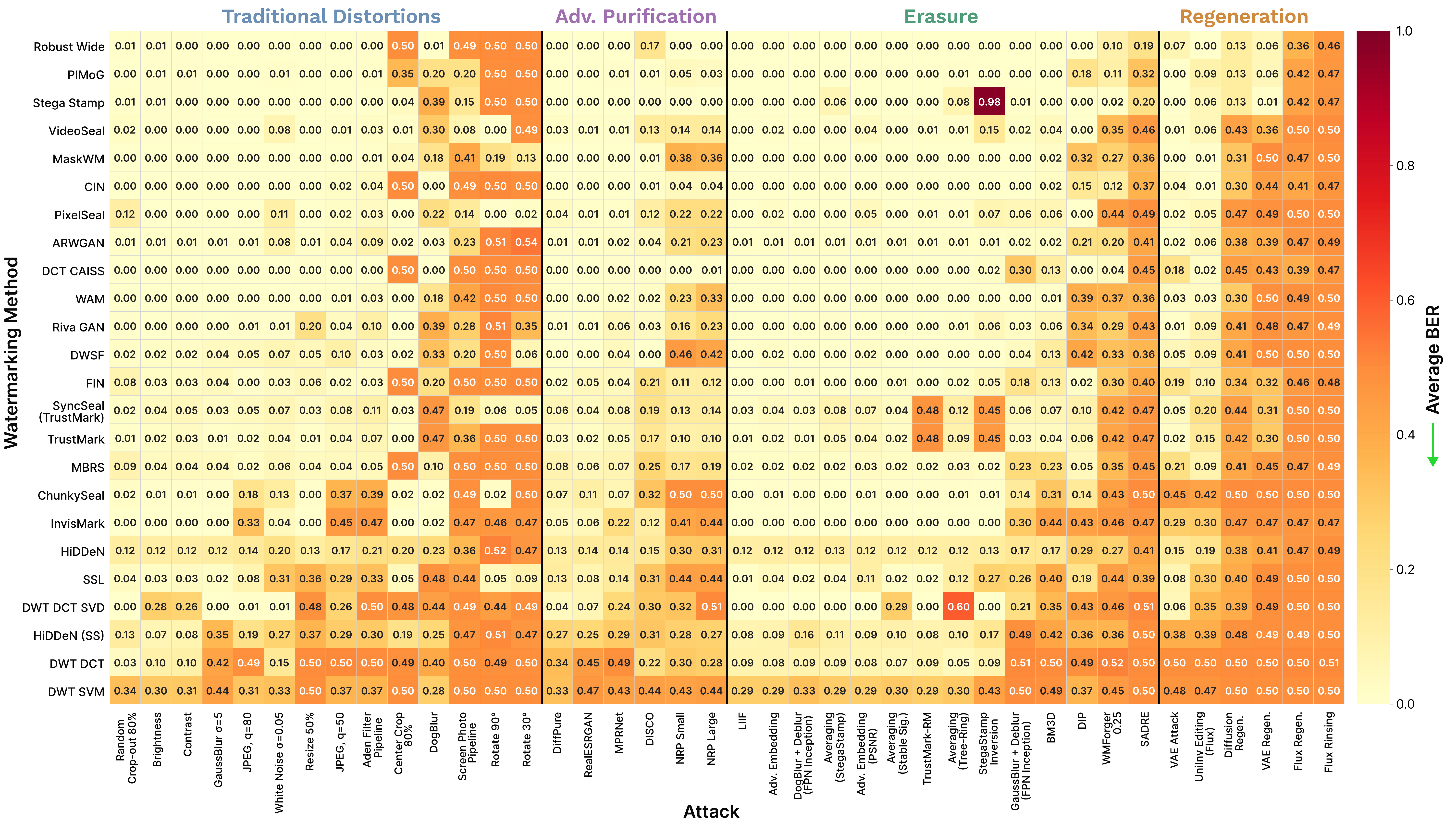}}
    \caption{Full robustness breakdown measured by Average BER ($\downarrow$) for all evaluated watermarking methods across all 40 attack configurations on the MS-COCO dataset. Each cell reports the mean BER over 1k images; values close to 0.5 indicate complete watermark erasure.}
    \Description{Heatmap of average bit error rate on the MS-COCO dataset, with the same
layout as the DiffusionDB breakdown: rows are post-hoc watermarking methods sorted by
increasing average bit error rate and columns are attack configurations grouped into
traditional distortions, adversarial purification, erasure and regeneration. The
pattern closely matches the DiffusionDB results, with near-zero values under
traditional distortions and purification for most methods, and values approaching 0.5
under regeneration, non-right-angle rotation and dedicated erasure attacks. Robust-Wide
and PIMoG have the lowest average bit error rate, and DWT SVM and DWT DCT the highest.}
\label{fig:heatmap_coco_ber}
\end{figure*}

\begin{figure*}[htbp]
    \centerline{\includegraphics[width=0.85\textwidth]{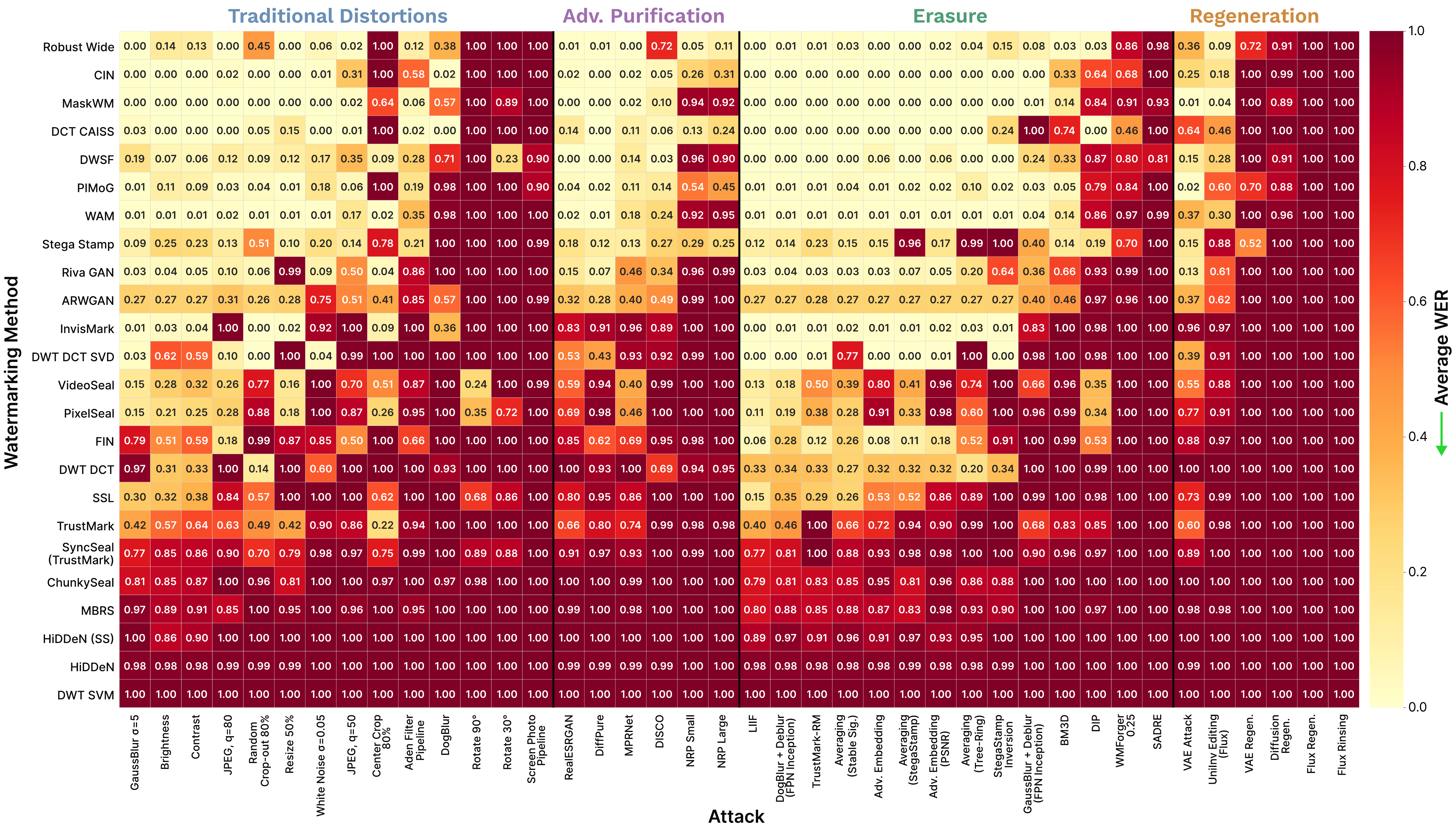}}
    \caption{Full robustness breakdown measured by Average WER ($\downarrow$) for all evaluated watermarking methods across all 40 attack configurations on the MS-COCO dataset. Each cell reports the mean WER over 1k images; values close to 1.0 indicate complete watermark erasure.}
    \Description{Heatmap of average word error rate on the MS-COCO dataset, where a value
of 1.0 means every message is lost. Rows are post-hoc watermarking methods sorted by
increasing average word error rate and columns are attack configurations grouped into
traditional distortions, adversarial purification, erasure and regeneration. Because a
single flipped bit invalidates the whole message, the heatmap is far more saturated than
the corresponding bit error rate figure: entire columns for regeneration attacks,
rotation and several erasure attacks are at 1.0 for nearly every method, and the HiDDeN,
HiDDeN (SS) and DWT SVM rows are close to 1.0 almost everywhere.}
\label{fig:heatmap_coco_wer}
\end{figure*}

Figures \ref{fig:heatmap_ber}, \ref{fig:heatmap_coco_ber} and \ref{fig:heatmap_coco_wer} present the full per-method, per-attack robustness breakdown across all evaluated post-hoc and generative watermarking methods on DiffusionDB and MS-COCO datasets, measured by Average BER ($\downarrow$, lower is better) and WER ($\downarrow$) respectively. TPR@0.01\%FPR results are presented in Fig. \ref{fig:main_res_diffdb} (b). The results confirm and extend the observations from the main paper. Regeneration-based attacks -- particularly Flux Regeneration, Flux Rinsing, and VAE Regeneration -- consistently produce the most severe watermark degradation, pushing BER toward 0.5 (chance level) and TPR toward zero for the vast majority of post-hoc methods, irrespective of their architecture family. Among traditional distortions, rotation through a non-right angle (30°) remains the single most damaging geometric transformation, collapsing robustness even for methods that handle JPEG compression, noise, and moderate cropping without difficulty. Methods with explicit spatial synchronization mechanisms, such as DWSF and DFT Circle, show notably higher resistance to rotation than architecturally similar peers without such components. Across all attack conditions, Robust-Wide stands out as the most consistently resilient post-hoc method, maintaining low BER and high TPR across both distortion and adversarial attack families -- a level of robustness that approaches that of generative approaches such as Gaussian Shading under non-regeneration settings. At the opposite end, classical frequency-domain methods (DWT DCT, DWT SVM) exhibit near-random extraction performance under the majority of attacks beyond mild JPEG compression, confirming their unsuitability for adversarially robust deployment.

\subsection{Cross-dataset and cross-resolution evaluation}
\label{sec:cross_dset}
\textbf{Setup.} To test whether our findings depend on the two datasets used in the main experiments, we re-evaluate all 24 multi-bit post-hoc watermarking methods on five datasets spanning an order of magnitude in pixel count: NIPS2017 ($299{\times}299$), DiffusionDB ($512{\times}512$), MS-COCO ($\sim$$640{\times}480$), KonIQ-10k ($1024{\times}768$) and DIV2K ($2040{\times}1300$). We use 1k images for DiffusionDB and MS-COCO and 400 images for the three additional datasets, and restrict the comparison to the 39 attacks that cover the identical set of watermarks on all five datasets, so that every reported average is taken over exactly the same conditions.\footnote{LIIF is the only excluded attack: it exceeds available GPU memory for part of the methods at the DIV2K resolution.} Built-in watermarks are omitted from this experiment, as they are tied to the output resolution of their underlying generative model and therefore have no dataset-resolution axis.

\begin{figure*}[htbp]
    \centerline{\includegraphics[width=0.85\textwidth]{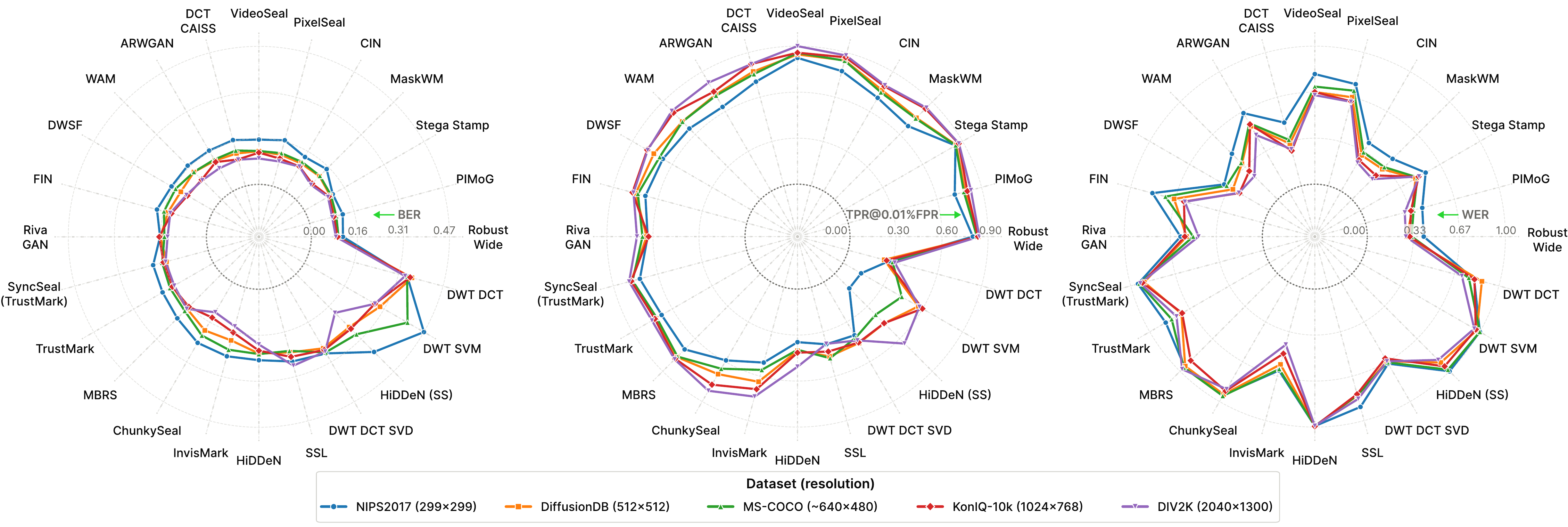}}
    \caption{Robustness scores of all tested multi-bit watermarks across samples from 5 datasets: DiffusionDB, MS-COCO, NIPS2017, KonIQ and DIV2K. Image resolution varies from 299x299 in NIPS2017 up to 2040x1300 in DIV2K. The scores are averaged across all attacks on each dataset.}
    \Description{Three radar charts, one for each robustness metric: average bit error rate,
average true positive rate at 0.01 percent false positive rate, and average word error
rate. In each chart, every axis corresponds to one of the 24 multi-bit post-hoc
watermarking methods and each coloured polygon to one of five datasets: NIPS2017 at 299
by 299 pixels, DiffusionDB at 512 by 512, MS-COCO at about 640 by 480, KonIQ-10k at
1024 by 768, and DIV2K at 2040 by 1300. The polygons are nearly concentric, showing
that the relative ranking of methods is essentially the same on all five datasets, while
the higher-resolution datasets sit consistently at slightly better values. The largest
spread between datasets occurs for DWT SVM and HiDDeN (SS), and the word error rate
chart is visibly more compressed than the other two.}
\label{fig:cross_dset_all}
\end{figure*}

\textbf{Ranking stability.} Figure~\ref{fig:cross_dset_all} shows the per-method scores for five tested datasets. The relative ordering of methods is essentially dataset-independent: the Spearman rank correlation with the MS-COCO ordering stays above $0.88$ for every dataset under all three metrics, and the qualitative conclusions of the main paper -- Robust Wide and StegaStamp as the most resilient post-hoc methods, the classical DWT DCT family close to chance -- hold unchanged on all five sets. What does shift is the absolute level: mean BER decreases monotonically with resolution across the dataset means, from $0.208$ on NIPS2017 to $0.147$ on DIV2K, with mean TPR@0.01\%FPR rising correspondingly from $0.590$ to $0.725$; 23 of 24 methods obtain a lower BER on DIV2K than on NIPS2017. The effect is driven by resolution rather than by image domain: DiffusionDB (synthetic, $512{\times}512$) and MS-COCO (photographic, $\sim$$640{\times}480$) sit adjacent in the middle of the range and differ by only $0.012$ BER.

\textbf{The effect of higher resolution.} The mechanism follows from the resolution-scaling protocol described in Section~\ref{sec:details}: every method embeds and extracts at its own fixed internal resolution, typically $224$--$512$ px. An attack is applied at the native resolution of the image, but the decoder downscales the result before extraction, and that downscaling acts as a low-pass filter on the attack's residual. Perturbations defined at a fixed pixel scale are therefore attenuated in proportion to the downscaling factor, which is exactly what we observe per attack: e.g., for JPEG ($q=50$) average BER drops from $0.20$ on NIPS2017 to $0.07$ on DIV2K, and for BM3D it falls from $0.24$ to $0.09$. Aggregated by family, adversarial purification loses roughly half of its effectiveness between the two extremes ($0.172$ BER on NIPS2017, $0.088$ on DIV2K), while regeneration attacks lose only $18\%$ ($0.371$ vs $0.306$): regeneration replaces image content through a generative model operating at its own working resolution instead of perturbing pixels, so it is largely immune to this averaging effect. The same argument explains why NIPS2017 is the hardest set rather than merely the smallest -- at $299{\times}299$ the image is \textit{below} the internal resolution of several methods and is upsampled rather than downsampled, so no averaging gain is available at all.

\textbf{Resolution sensitivity.} The per-method spread across datasets is modest on average -- $0.066$ BER between the best and worst dataset, against a between-method range of $0.35$ -- but it is highly uneven, from $0.019$ for StegaStamp to $0.193$ for DWT SVM. The spread tracks where a method sits on its own robustness--distortion curve rather than its architecture family. Methods near either end of that curve are flat: StegaStamp and Robust Wide stay in the $0.084$--$0.112$ BER band on every dataset because they are already close to the achievable robustness ceiling, while DWT DCT and DWT DCT SVD stay at $0.34$ and $0.27$ because they are already close to chance level under most attacks, and a change in effective attack strength moves neither. The largest spreads belong to methods in the steep middle of the curve, and these are disproportionately the high-capacity and high-imperceptibility ones -- DWT SVM ($512$ bits, spread $0.193$), HiDDeN~(SS) ($0.189$), ChunkySeal ($1024$ bits, $0.120$) and InvisMark (highest PSNR of all evaluated methods, $0.105$) -- where the per-bit decision margin is thinnest and a small change in the effective noise level flips a large number of bits at once. WER metric in Figure~\ref{fig:cross_dset_all} illustrates the complementary saturation effect: because a single flipped bit invalidates the message, WER compresses the methods that BER separates, and HiDDeN, whose WER is $0.991$--$0.994$ on all five datasets, becomes indistinguishable across resolutions. The practical implication is that benchmark conclusions transfer across resolution and image domain, but \textit{absolute} robustness figures for high-capacity watermarks may vary across different image sizes, and should be reported together with the evaluation resolution.

\subsection{Metrics for perceptual and semantic changes}
\label{sec:metrics}

\begin{figure*}[htbp]
    \centerline{\includegraphics[width=0.85\textwidth]{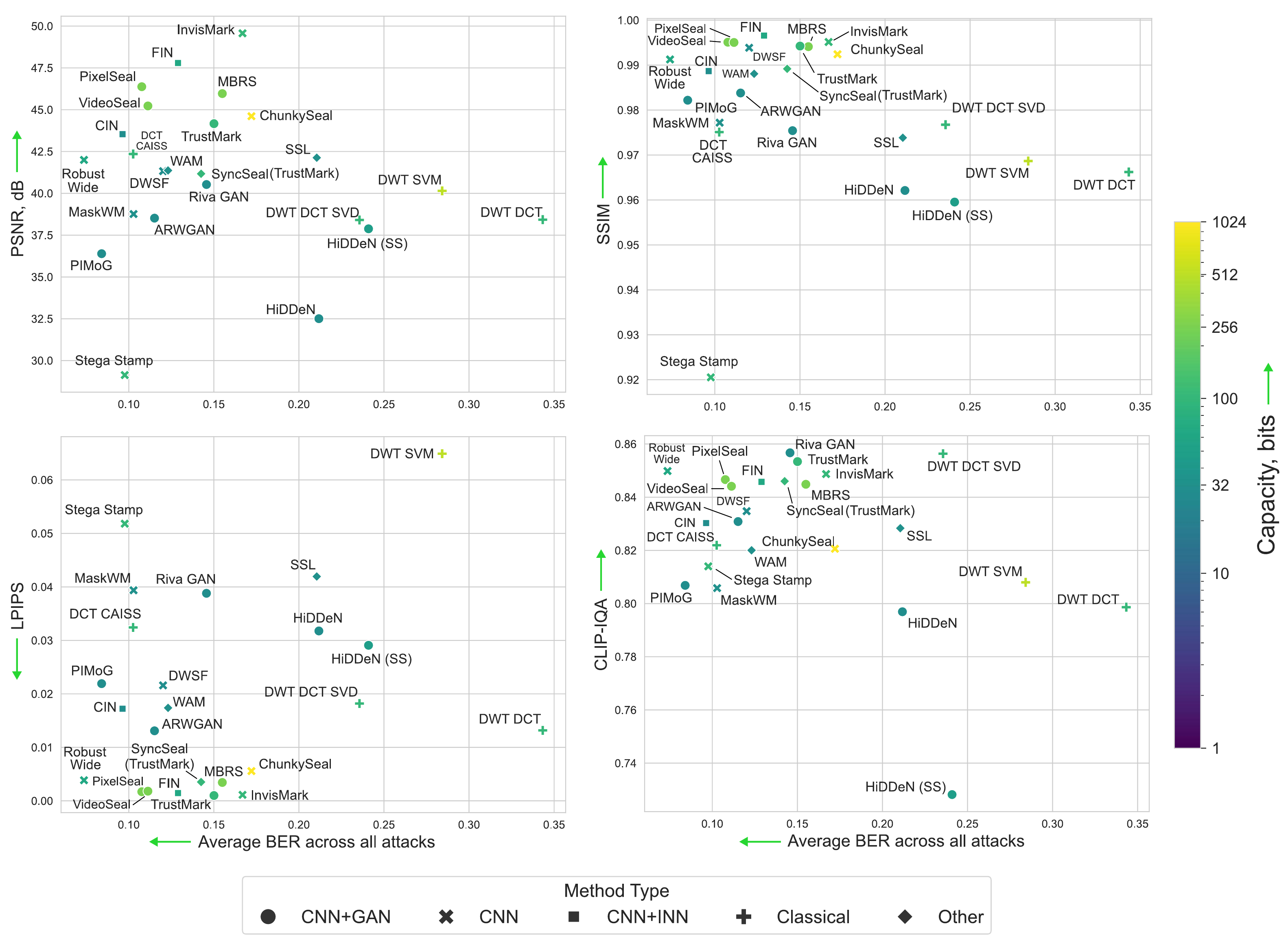}}
    \caption{Quality–robustness tradeoff for post-hoc watermarking methods evaluated under four imperceptibility metrics: PSNR (dB, $\uparrow$), SSIM ($\uparrow$), LPIPS ($\downarrow$), and CLIP-IQA ($\uparrow$). Robustness is reported as average BER across all attacks ($\downarrow$). Color encodes embedding capacity in bits. While the overall rankings are broadly consistent across metrics, notable divergences between full-reference (PSNR, SSIM, LPIPS) and no-reference (CLIP-IQA) measures reveal that different metrics capture complementary aspects of watermark visibility.}
    \Description{Four scatter panels sharing an x-axis of average bit error rate across all
attacks, where lower is better, and plotting on the y-axis, respectively, PSNR in
decibels, SSIM, LPIPS and CLIP-IQA, for post-hoc watermarking methods, with colour
encoding embedding capacity in bits. Rankings are broadly consistent across the four
metrics: methods with low bit error rates also tend to score well on perceptual quality.
InvisMark attains the highest PSNR of any method at about 50 decibels, and StegaStamp is
worst on PSNR, SSIM and LPIPS. Two divergences stand out: DWT SVM has one of the worst
LPIPS values despite an unremarkable SSIM, and HiDDeN (SS) has the lowest CLIP-IQA of
any method at about 0.73 despite mid-range PSNR and SSIM.}
\label{fig:watermarks_different_metrics}
\end{figure*}

\textbf{Watermark visibility evaluation with different metrics}. Figure \ref{fig:watermarks_different_metrics} extends the capacity–quality–robustness analysis from the main paper by reporting imperceptibility across four complementary metrics: PSNR and SSIM (classical full-reference), LPIPS (learned full-reference), and CLIP-IQA (learned no-reference). Across all four metrics, a broadly consistent picture emerges: methods with low average BER tend to also achieve better perceptual quality, confirming that robustness and imperceptibility are not inherently in tension when methods are carefully designed. InvisMark is a particularly notable outlier in this regard, achieving the highest PSNR of any evaluated method ($\sim$50 dB) while maintaining competitive robustness. StegaStamp sits at the opposite extreme across all four metrics -- lowest PSNR, lowest SSIM, and highest LPIPS -- reflecting its aggressive perturbation strategy, which trades imperceptibility for robustness. However, the metrics diverge in informative ways when comparing specific methods. DWT SVM, a classical multi-bit method with high capacity and low robustness, registers one of the worst LPIPS scores despite appearing relatively benign under SSIM; this discrepancy likely reflects LPIPS's sensitivity to fine-grained texture distortions that SSIM averages over. Conversely, HiDDeN (SS) achieves moderate PSNR and SSIM but drops sharply in CLIP-IQA ($\sim$0.73) -- the lowest of any method -- suggesting that its embedding introduces perceptual artifacts that are not well captured by pixel-level full-reference measures but are detected by a learned no-reference predictor operating on absolute image quality. Taken together, these results reinforce that no single metric fully characterizes watermark visibility: full-reference metrics capture fidelity to the original, while no-reference metrics reflect the absolute perceptual plausibility of the watermarked image — both perspectives are necessary for a complete evaluation.

\begin{figure*}[htbp]
    \centerline{\includegraphics[width=0.99\textwidth]{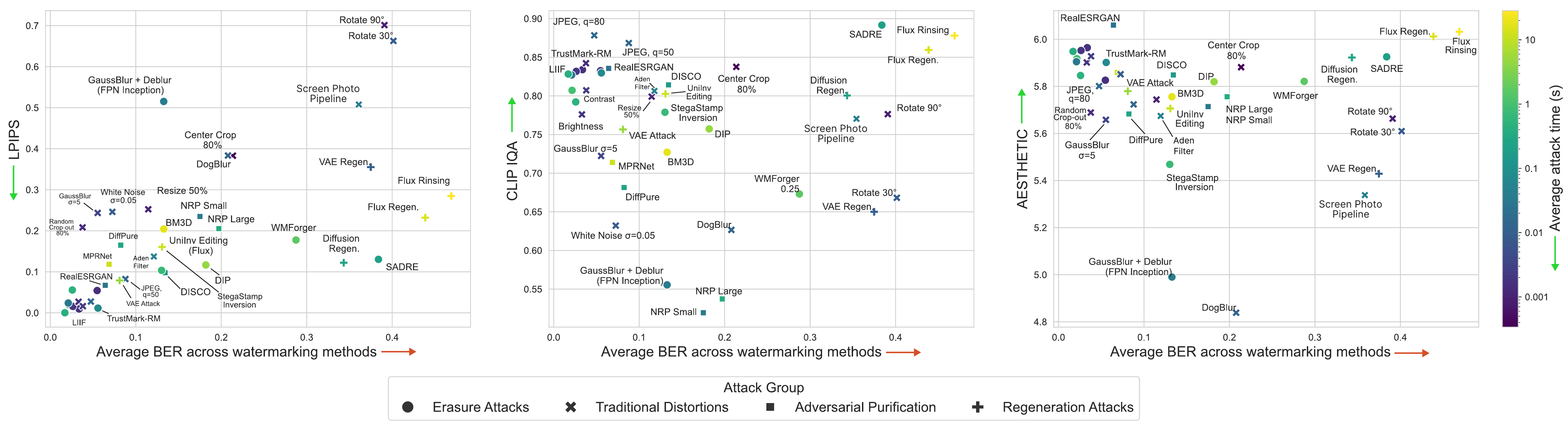}}
    \caption{Perceptual quality–adversarial effectiveness tradeoff for all evaluated attacks, measured under three quality criteria: LPIPS ($\downarrow$), CLIP-IQA ($\uparrow$), and Aesthetic score ($\uparrow$). Attack effectiveness is reported as average BER across all watermarking methods ($\uparrow$). Color encodes average attack runtime. Regeneration-based attacks (crosses) achieve high BER with strong no-reference quality scores, indicating that they replace rather than degrade the image. Geometric distortions incur high LPIPS but mostly preserve semantic quality.}
    \Description{Three scatter panels sharing an x-axis of average bit error rate across all
watermarking methods, where higher means a more effective attack, and plotting LPIPS,
CLIP-IQA and aesthetic score on the y-axes for every evaluated attack. Marker shape
encodes attack group and colour encodes average runtime. Regeneration attacks, in
particular Flux Regeneration and Flux Rinsing, reach the highest bit error rates with
elevated LPIPS but also the highest CLIP-IQA and aesthetic scores, indicating that they
re-synthesise rather than degrade the image. Rotations by 30 and 90 degrees show high
LPIPS while leaving the no-reference scores largely intact. GaussBlur with deblurring
and DogBlur score poorly on both effectiveness and quality, while WMForger and
TrustMark-RM lie on the favourable frontier of all three panels.}
\label{fig:attacks_different_metrics}
\end{figure*}

\textbf{Attacked image clarity evaluation}. Figure \ref{fig:attacks_different_metrics} evaluates attack quality from three complementary perspectives — LPIPS, CLIP-IQA, and an aesthetic predictor — plotted against attack effectiveness (average BER). A striking asymmetry is immediately apparent between full-reference and no-reference quality assessments for regeneration-based attacks. Flux Regeneration and Flux Rinsing achieve among the highest BER values in the benchmark while simultaneously scoring at or near the top of both CLIP-IQA and Aesthetic measures; this indicates that these attacks do not merely corrupt images but actively re-synthesize them at high perceptual quality, producing outputs that a no-reference evaluator scores as visually superior to the original watermarked image. In contrast, their LPIPS values are elevated, correctly capturing the significant pixel-level departure from the source — a distinction that is invisible to no-reference metrics. A similar but less extreme pattern holds for SADRE and Diffusion Regeneration. Geometric distortions, particularly rotation (30° and 90°), occupy a distinct region: they rank among the strongest traditional attacks in terms of BER and show high LPIPS due to spatial misalignment, yet their CLIP-IQA and Aesthetic scores remain largely intact, since the semantic content of the image is undamaged. This confirms that full-reference metrics overstate the perceptual damage of geometric transforms relative to their semantic impact. At the other extreme, GaussBlur+Deblur (FPN Inception) and DogBlur achieve poor BER while also scoring low on both CLIP-IQA and Aesthetic — they are simultaneously ineffective as attacks and destructive to image quality, making them unfavorable from both perspectives. Among dedicated erasure attacks, WMForger and TrustMark-RM dominate the Pareto front across all three quality measures, confirming that purpose-built removal methods can combine high watermark disruption with minimal perceptual footprint regardless of which quality metric is used to evaluate them.

\subsection{Attack hyperparameter variation}
\label{sec:attack_hyp}

Figure \ref{fig:atk_variation} presents BER–distortion curves for six selected attacks across individual watermarking methods on the COCO dataset, where each attack's primary hyperparameter is varied: learning rate for WMForger, $\epsilon$ ($L_{\inf}$ perturbation budget) for Adversarial Embedding, quality factor for JPEG compression, std for Gaussian noise, and kernel size for Gaussian blur. The per-method curves reveal qualitatively different regimes across the five attack types. WMForger produces the most definitive pattern: nearly all tested methods exhibit a sharp BER transition concentrated within a narrow LPIPS window of approximately 0.12–0.20, after which extraction collapses to chance level. This uniformity suggests that WMForger's optimization finds a common vulnerability threshold across architecturally diverse methods. The single clear exception is DCT CAISS, which remains highly resistant throughout the entire evaluated learning rate range, reaching only $\sim$0.35 BER at LPIPS $\sim$0.35 -- far below the saturation level reached by all other methods. For Adversarial Embedding (both CLIP and ResNet variants), curves are considerably more spread out and gradual, indicating that gradient-based perturbation exploits method-specific decoder properties rather than a shared threshold, with HiDDeN (SS) among the most vulnerable and DCT CAISS again the most resistant. In the JPEG panel, InvisMark shows a disproportionately steep early BER increase under light compression, revealing a specific sensitivity to quantization artifacts that its training augmentation does not fully compensate for. For the Gaussian blur, DCT CAISS displays a pronounced threshold effect — BER remains near zero across a wide range of kernel sizes before abruptly jumping once blur exceeds a critical scale, consistent with its frequency-domain embedding concentrating energy in a band that broad blur eventually destroys. Noise curves are the smoothest overall, with most methods degrading gradually and DCT CAISS and WAM showing exceptional robustness even at very large noise standard deviations.

\begin{figure}[tbp]
    \centerline{\includegraphics[width=0.9\columnwidth]{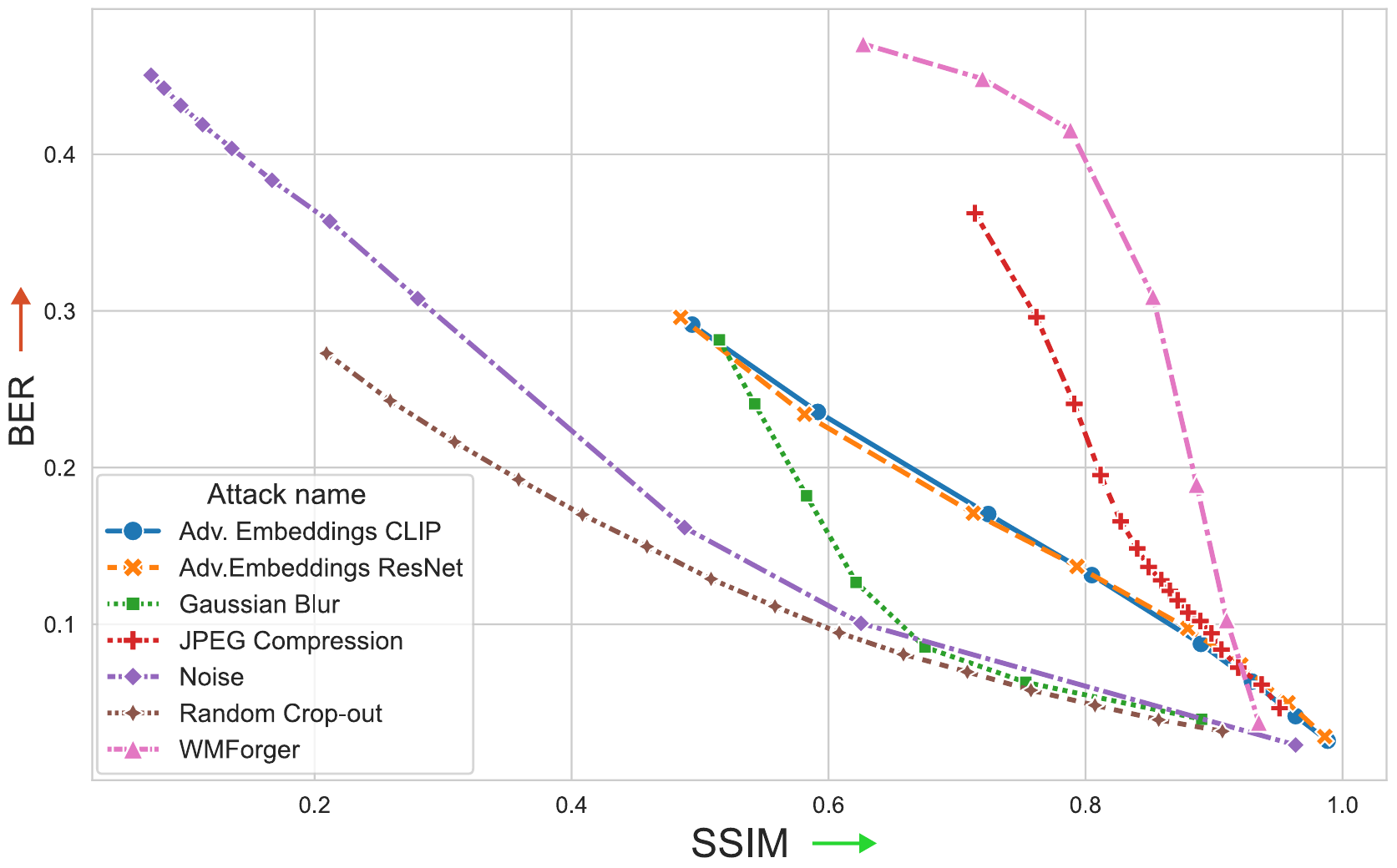}}
    \caption{BER–SSIM tradeoff curves averaged across watermarks for each evaluated attack under hyperparameter variation on the COCO dataset. Higher SSIM indicates less image distortion; higher BER indicates more effective watermark removal.}
    \Description{Line plot of average bit error rate on the y-axis from 0 to about 0.45
against SSIM on the x-axis from 0.1 to 1.0, averaged over all watermarking methods on
the MS-COCO dataset, with one curve per attack whose main hyperparameter is swept: two
Adversarial Embedding variants, Gaussian blur, JPEG compression, Gaussian noise, random
crop-out and WMForger. Every curve rises as SSIM falls, so the most efficient attacks
lie towards the upper right. WMForger dominates, reaching high bit error rates at SSIM
above 0.8, followed by JPEG compression and the two Adversarial Embedding variants,
while Gaussian blur, Gaussian noise and random crop-out need far more image distortion
to reach comparable bit error rates.}
\label{fig:atk_variation_avg}
\end{figure}

\begin{figure*}[htbp]
    \centerline{\includegraphics[width=0.9\textwidth]{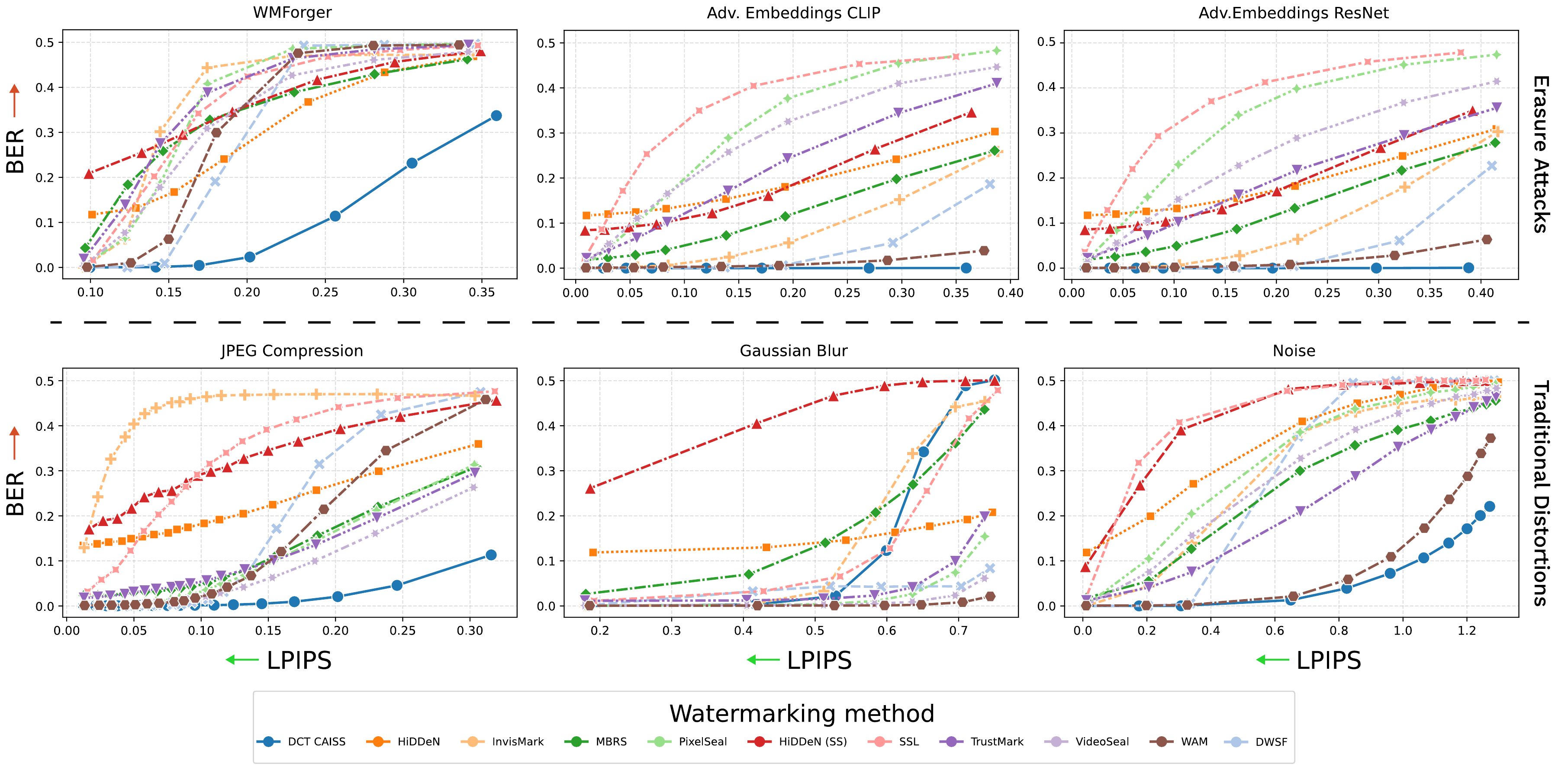}}
    \caption{BER–distortion curves under hyperparameter variation for six attacks, shown per watermarking method on the COCO dataset. Each panel varies the attack's primary hyperparameter: learning rate (WMForger), $L_{\inf}$ budget $\epsilon$ (Adversarial Embedding), quality factor (JPEG), std (Gaussian noise), and kernel size (Gaussian blur). Distortion is measured by LPIPS ($\downarrow$).}
    \Description{Six panels in two rows of three, each plotting bit error rate on the y-axis
from 0 to 0.5 against LPIPS distortion on the x-axis, with one curve per watermarking
method, as the primary hyperparameter of an attack is varied on the MS-COCO dataset. The
top row shows erasure attacks: WMForger with varying learning rate and Adversarial
Embedding with CLIP and with ResNet backbones with varying perturbation budget. The
bottom row shows traditional distortions: JPEG compression with varying quality factor,
Gaussian blur with varying kernel size, and Gaussian noise with varying standard
deviation. Under WMForger nearly every method transitions sharply to chance-level bit
error rate within a narrow LPIPS window of roughly 0.12 to 0.20, whereas the Adversarial
Embedding curves are gradual and widely spread. DCT CAISS stays far below all other
methods in most panels and shows a delayed threshold jump under Gaussian blur, while
InvisMark degrades unusually quickly under light JPEG compression.}
\label{fig:atk_variation}
\end{figure*}

\begin{figure*}[htbp]
    \centerline{\includegraphics[width=0.8\textwidth]{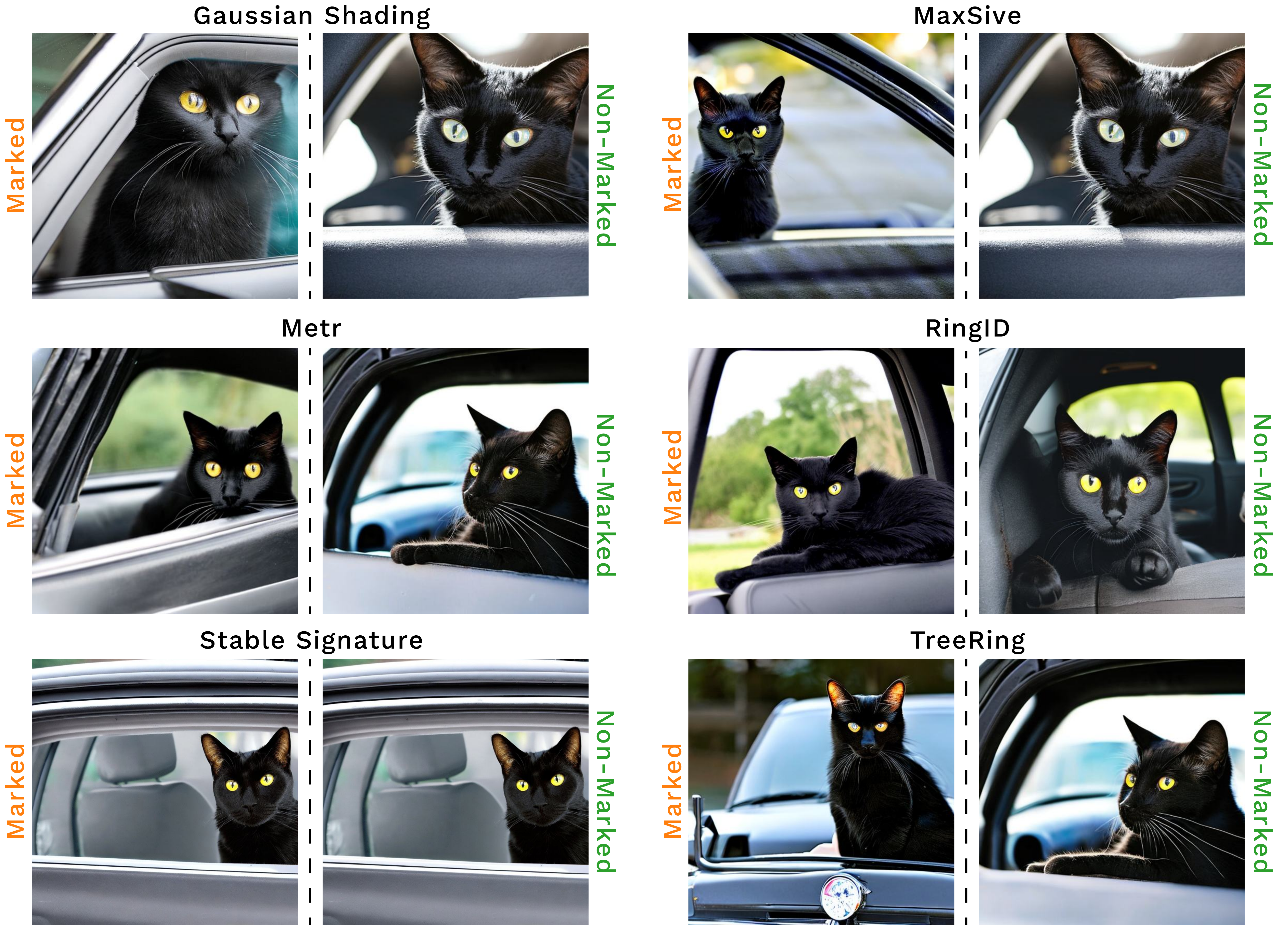}}
    \caption{Examples of images generated with different built-in watermarking methods compared to the ones generated by the same underlying generative model with the same prompt and seed without watermark insertion. The prompt used for all presented generations: "A black cat sits in a car and looks out."}
    \Description{Six pairs of generated photographs of a black cat looking out of a car
window, one pair for each built-in watermarking method: Gaussian Shading, MaXsive, METR,
RingID, Stable Signature and TreeRing. In each pair the left image is watermarked and
the right image was produced by the same generative model with the same prompt and seed
without watermark insertion. Watermarked and non-watermarked images are of comparable
visual quality with no visible artefacts. For the noise-space methods the two images of
a pair differ in composition and framing rather than in fidelity, whereas Stable
Signature, which watermarks through decoder fine-tuning, preserves the composition
almost exactly.}
\label{fig:examples_gen_wms}
\end{figure*}

Figure \ref{fig:atk_variation_avg} aggregates the per-watermark curves into mean BER vs SSIM tradeoff curves per attack, enabling a direct cross-attack comparison of distortion efficiency. Note that SSIM is used as the distortion axis here (higher SSIM indicates less image change), so the ideal attacker occupies the upper-right region — high BER at high SSIM. WMForger remains the most cost-efficient overall, with JPEG and Adversarial Embedding (with both CLIP and ResNet backbones) attacks following it.

\subsection{Frequency patterns of different watermarks}
\label{sec:freq_examples}

\begin{figure*}[htbp]
    \centerline{\includegraphics[width=0.85\textwidth]{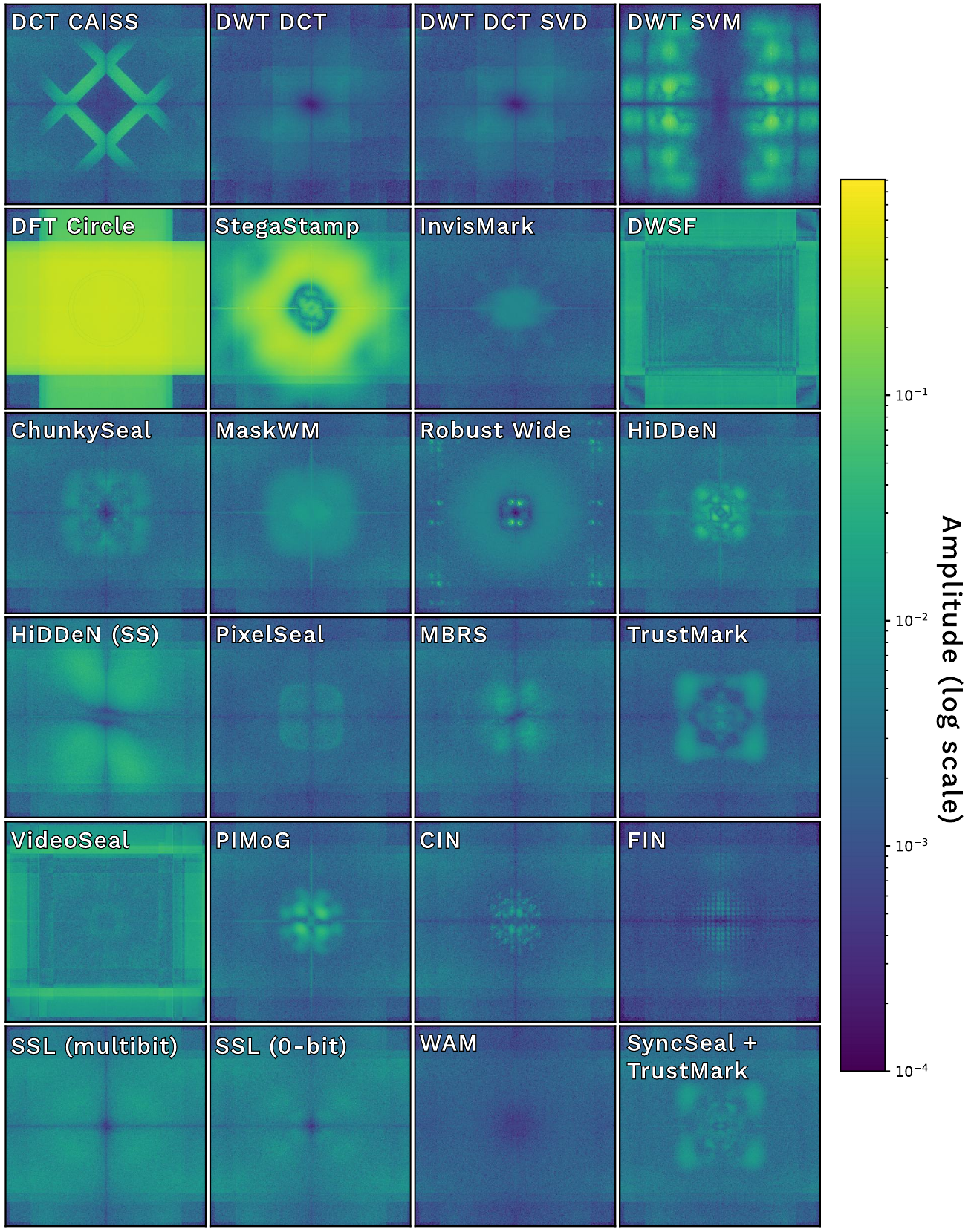}}
    \caption{Frequency-domain signatures of post-hoc watermarking methods, visualized as the average FFT spectrum difference between watermarked and source images on the COCO subset. Amplitude is shown on a logarithmic scale.}
    \Description{A grid of 24 frequency-domain maps, one per post-hoc watermarking method,
showing the average FFT amplitude difference between watermarked and source images on
the MS-COCO subset, on a logarithmic colour scale spanning four orders of magnitude.
Classical methods produce highly structured signatures: a prominent X-shaped cross for
DCT CAISS, a regular grid-like lattice for DWT SVM, a bright square frame at the image
boundary for DWSF and VideoSeal, and a uniformly bright field for DFT Circle. Among
learned methods, StegaStamp shows a strong ring pattern at mid-to-high frequencies and
TrustMark a structured square motif, while FIN and CIN show faint fine-grained dot
patterns. InvisMark, MaskWM, WAM and PixelSeal are almost featureless, indicating
spectrally diffuse embeddings.}
\label{fig:more_freq_examples}
\end{figure*}

Figure \ref{fig:more_freq_examples} visualizes the average FFT spectrum difference between watermarked and source images for each post-hoc method, revealing the frequency-domain footprint of each embedding strategy. Classical methods produce structured and recognizable signatures: DCT CAISS displays a prominent X-shaped pattern with strong amplitude spread across mid-frequencies, a consequence of its spread-spectrum design. DWT SVM similarly shows a regular grid-like lattice, while DWSF's distinctive square-frame pattern at the image boundaries reflects its synchronization mechanism, directly linking its architectural choice to its geometric robustness. Among deep learning methods, StegaStamp produces a clearly visible ring pattern with high amplitudes at mid-to-high frequencies, consistent with its high perceptual cost. In contrast, methods such as InvisMark, MaskWM, WAM and PixelSeal leave minimal or near-invisible frequency signatures, indicating that their embeddings are more broadly distributed and spectrally diffuse -- a property that might underlie their stronger imperceptibility scores. FIN and CIN exhibit faint but structured fine-grained dot patterns, reflecting the invertible network architectures that distribute perturbations across many frequency bands simultaneously. The structural diversity visible across these spectra also might be particularly relevant to the asymmetric re-embedding results reported in Section \ref{sec:reembed_attacks}: methods whose frequency footprints occupy largely non-overlapping regions (e.g., DCT CAISS and StegaStamp) tend to interact less destructively when superimposed, while methods sharing similar spectral support are more likely to mutually interfere.

\section{Limitations and Future Work}
\label{sec:limitations}
This work presents a large-scale and systematic evaluation of invisible image watermarking methods; however, several limitations should be acknowledged. First, the benchmark reflects the state of the field at the time of submission and does not include future methods that may introduce stronger robustness guarantees, particularly as generative and key-based approaches continue to evolve. As a result, the reported findings should be interpreted as a snapshot of current capabilities rather than a definitive assessment of long-term progress. At the same time, WARP is designed as a modular and extensible framework, allowing straightforward integration of new watermarking techniques and attack strategies, enabling future updates under consistent evaluation protocols.

Second, the scope of this work is restricted to image watermarking. While this domain represents a primary use case, the problem of robust and traceable watermarking extends to other media types, including video, audio, and text, each with distinct perceptual constraints and attack surfaces. These modalities are not considered in the current benchmark, limiting the generality of conclusions across multimodal settings.

Third, although the benchmark includes a diverse set of 34 attack scenarios spanning distortions, adversarial perturbations, purification, and regeneration, it does not exhaust the full space of possible removal strategies. In particular, fully adaptive attacks tailored to specific watermarking methods, as well as complex real-world transformation pipelines, may lead to different robustness outcomes. Additionally, the evaluation focuses on perceptual quality, watermark readability, and robustness, while other relevant factors such as formal security guarantees are not explicitly addressed.

Finally, the proposed re-embedding evaluation protocol captures an important but previously underexplored threat model; however, it relies on specific assumptions about attacker capabilities and access conditions. Alternative threat settings may yield different results and require further investigation.

Future work will focus on extending WARP along several directions. A key priority is continuous integration of newly proposed watermarking methods and attack strategies to maintain an up-to-date and longitudinal benchmark. Another important direction is the expansion toward multimodal watermarking, enabling unified evaluation across images, video, audio, and text, and facilitating analysis of modality-specific and cross-modal robustness. Further efforts will include the development of more adaptive and method-aware attack models, as well as the formalization of standardized threat models to improve comparability across studies. Finally, incorporating additional evaluation criteria, including efficiency, scalability, and robustness under real-world processing pipelines, will be essential for bridging the gap between benchmark performance and practical deployment.

\begin{table*}[h]
\caption{All attacking methods used in our study alongside their brief descriptions.}
\centering
\resizebox{1.8\columnwidth}{!}{

\begin{tabular}{lccp{10cm}}
\toprule

Attack & Year & Type & Brief description\\
\midrule

Brightness & - & Distortion & Changes image intensity levels, making embedded patterns harder to preserve or detect \\
Center Crop & - & Distortion & Removes outer regions of an image, potentially discarding embedded information \\
Color Inversion & - & Distortion & Flips pixel values, altering image patterns and disrupting embedded information \\
Contrast & - & Distortion & Rescales intensity differences, altering pixel relationships and distorting image patterns \\
Gaussian Blur & - & Distortion & Smooths the image using a Gaussian filter, reducing detail and weakening fine patterns \\
JPEG & - & Distortion & Lossy quantization that removes high-frequency details and introduces compression artifacts \\
White Noise & - & Distortion & Random pixel intensity perturbations that degrade image quality \\
Pixel Shift \cite{shamshadFirstPlaceSolutionNeurIPS2025} & 2025 & Distortion & Translates the image horizontally by shifting the image pixels  \\
Random Cropout & - & Distortion & Random removal of image regions causing partial content loss \\
Resize & - & Distortion & Scaling image dimensions, leading to blurring or detail loss \\
Rotate & - & Distortion & Angular transformation causing geometric distortion and potential cropping \\
Aden Filter Pipeline & - & Distortion & A pipeline that chains 3 different transformations together: Gaussian~noise $\rightarrow$ Aden filter $\rightarrow$ JPEG,~q=50. Simulates social media processing of a user image with a popular color filter~\cite{pilgram} and compression.  \\
Screen Photo Pipeline & - & Distortion & A pipeline that chains 3 different transformations together: $0.7\times$downscale~$\rightarrow$ JPEG,~q=60 $\rightarrow$~Perspective transform with slight angle~$\rightarrow$ Gaussian~noise $\rightarrow$ JPEG,~q=50. Simulates a slightly angled photo of a screen with a given image.  \\
Adversarial Embedding \cite{an2024waves} & 2024 & Erasure & Added imperceptible perturbations to an image to mislead the watermark detector and disrupt watermark extraction \\
Adversarial Embedding (PSNR) & - & Erasure & Added imperceptible perturbations to an image in order to mislead the watermark detector using PSNR metric to constrain the perturbation \\
Averaging \cite{yangSteganalysisDigitalWatermarking2024} & 2024 & Erasure & Averaging attack estimates the watermark by averaging multiple samples, canceling content and exposing the hidden pattern \\
Blur Deblur \cite{kupyn2019deblurgan} & 2019 & Erasure & Applies intentional blurring followed by restoration (deblurring) to disrupt and attenuate embedded watermark signals.\\
BM3D \cite{makinen2020collaborative} & 2020 & Erasure & Exploits transform-domain denoising to estimate and suppress correlated noise components, potentially weakening or removing embedded watermark signals\\
DIP \cite{liangBaselineMethodRemoving2025} & 2025 & Erasure & Reconstructed the image via an untrained neural network, implicitly denoising and removing watermark patterns \\
DoGBlur + Deblur \cite{kupyn2019deblurgan} & 2019 & Erasure & Applies intentional blurring using Difference of Gaussians filter to eliminate specific frequencies affected by the watermark followed by restoration (deblurring). \\
SADRE \cite{alam2025saliency} & 2025 & Erasure & Uses saliency-guided diffusion reconstruction with targeted noise injection to remove watermarks \\
SEMAttack \cite{mullerBlackBoxForgeryAttacks2025} & 2024 & Erasure & Performs single-image watermark forgery or removal using latent manipulations in unrelated diffusion models \\
Stegastamp Inversion \cite{serzhenko2025watermark} & 2025 & Erasure & Overwrote the watermark by decoding the embedded message, inverting its bits, and re-embedding it, thereby corrupting the original watermark signal \\
TrustMark-RM \cite{bui2023trustmark} & 2023 & Erasure & Denoising-based network designed to remove TrustMark watermarks and suppress watermark artifacts \\
WMForger \cite{souvcek2025transferable} & 2025 & Erasure & Optimizes input images via a preference model to remove or forge watermarks using a single reference \\
DiffPure \cite{nie2022diffusion} & 2022 & Purification & Uses diffusion models to remove adversarial perturbations via forward-noise diffusion and reverse generative reconstruction \\
LIIF \cite{chen2021learning} & 2020 & Purification & Resampled the image via a continuous implicit representation, altering local structures and potentially degrading embedded watermark patterns. \\
DISCO \cite{ho2022disco} & 2022 & Purification & Removes adversarial perturbations via localized manifold projections using per-pixel deep features \\
MPRNet \cite{zamir2021multi} & 2021 & Purification & Combines contextual and high-resolution features with per-pixel adaptive attention, enabling progressive denoising, deblurring, and deraining \\
NRP \cite{naseer2020self} & 2020 & Purification & Self-supervised adversarial training in the input space \\
REALSRGAN \cite{wang2021real} & 2021 & Purification & Extends ESRGAN to restore real-world low-resolution images with complex degradations \\
Diffusion Regeneration \cite{zhaoInvisibleImageWatermarks2024} & 2023 & Regeneration & Adds random noise and reconstructs images using generative models to remove invisible watermarks \\
FLUX Regeneration \cite{flux2024} & 2024 & Regeneration & Diffusion Regeneration using FLUX model \\
FLUX Rinsing \cite{flux2024} & 2024 & Regeneration & Diffusion Regeneration using FLUX model sequentially applied several times \\
FLUX VAE Regeneration \cite{flux2024} & 2024 & Regeneration & Encodes image with the VAE encoder from the FLUX model, adds noise and decodes with the decoder \\
Image Editing InstructPix2Pix \cite{brooks2023instructpix2pix} & 2022 & Regeneration & Uses a conditional diffusion model trained on synthetic image-edit instruction pairs to perform image edits \\
UniInv Image Editing (FLUX) \cite{jiao2026unieditflow} & 2026 & Regeneration & Uses a generative flow matching model with Uni-Inv sampling pipeline to edit the image \\
VAE Regeneration \cite{balle2018variational} & 2023 & Regeneration & End-to-end trainable VAE with hyperprior for image compression \\
\bottomrule

\end{tabular}

}
\label{tab:attacks_table}
\end{table*}



\end{document}